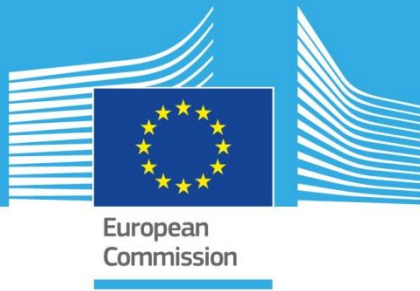

# JRC CONFERENCE AND WORKSHOP REPORTS

# Assessing the impact of machine intelligence on human behaviour: an interdisciplinary endeavour

*Proceedings of 1st HUMAINT workshop, Barcelona, Spain, March 5-6, 2018*

*Centre for Advanced Studies*

Gómez, Emilia (editor)

2018

This publication is a Conference and Workshop report by the Joint Research Centre (JRC), the European Commission's science and knowledge service. It aims to provide evidence-based scientific support to the European policymaking process. The scientific output expressed does not imply a policy position of the European Commission. Neither the European Commission nor any person acting on behalf of the Commission is responsible for the use that might be made of this publication.

**Contact information**
Name: Emilia Gómez
Email: Emilia.GOMEZ-GUTIERREZ@ec.europa.eu

**JRC Science Hub**
https://ec.europa.eu/jrc

JRC111773

Luxembourg: Publications Office of the European Union, Seville: European Commission, 2018

# Contents

# Acknowledgements

This workshop was organized by the Centre for Advanced Studies, Joint Research Centre with the local support of the Department of Information and Communication Technologies, Universitat Pompeu Fabra.

***Authors (editor plus authors in alphabetical order)***

Emilia Gómez (editor)

Carlos Castillo

Vicky Charisi

Verónica Dahl

Gustavo Deco

Blagoj Delipetrev

Nicole Dewandre

Miguel Ángel González-Ballester

Fabien Gouyon

José Hernández-Orallo

Perfecto Herrera

Anders Jonsson

Ansgar Koene

Martha Larson

Ramón López de Mántaras

Bertin Martens

Marius Miron

Rubén Moreno-Bote

Nuria Oliver

Antonio Puertas Gallardo

Heike Schweitzer

Nuria Sebastian

Xavier Serra

Joan Serrà

Songül Tolan

Karina Vold

## Abstract

This document contains the outcome of the first Human behaviour and machine intelligence (HUMAINT) workshop that took place 5-6 March 2018 in Barcelona, Spain. The workshop was organized in the context of a new research programme at the Centre for Advanced Studies, Joint Research Centre of the European Commission, which focuses on studying the potential impact of artificial intelligence on human behaviour.

The workshop gathered an interdisciplinary group of experts to establish the state of the art research in the field and a list of future research challenges to be addressed on the topic of human and machine intelligence, algorithm's potential impact on human cognitive capabilities and decision making, and evaluation and regulation needs.

The document is made of short position statements and identification of challenges provided by each expert, and incorporates the result of the discussions carried out during the workshop. In the conclusion section, we provide a list of emerging research topics and strategies to be addressed in the near future.

# 1   Human behaviour and machine intelligence in the digital transformation project HUMAINT: objectives and workshop

Emilia Gómez

Centre for Advanced Studies, Joint Research Centre, European Commission
Universitat Pompeu Fabra

Over the last few years, thanks to an increase in data availability and computing power, deep learning techniques have been applied to different research problems related to computer vision, natural language processing, music processing or bioinformatics. Some of these models are said to surpass human-level performance (e.g. image recognition (He et al., 2015) and model highly abstract human concepts such as emotion (Kim et al., 2013) or culture. The practical exploitation of such algorithms brings up a discussion on the impact of these algorithms into the ways human behave:

- On one side, machine intelligence provides **cognitive assistance and complement humans** to interpret data more efficiently and discover hidden knowledge in large data resources.
- On the other side, these algorithms may also **affect** the way we perform some **cognitive tasks and thus affect autonomy and decision making**. This is especially relevant when algorithms perform tasks in a high level of abstraction and when they may contradict and influence human interpretations.

The goal of the *Human behaviour and Machine Intelligence* (HUMAINT) project, carried out at the Centre for Advanced Studies, Joint Research Centre of the European Commission, is to (1) provide a scientific understanding of machine vs human intelligence; (2) analyse the influence of current algorithms on human behaviour and (3) investigate to what extent these findings should influence the European regulatory framework.

This document summarizes the results of the first HUMAINT workshop, which took place in Barcelona on 5-6 March 2018, and brought together key researchers from complementary disciplines and backgrounds. The workshop was defined with two main goals:

1. Build an interdisciplinary roadmap on human vs machine intelligence, potential algorithm's impact on human cognitive capabilities and decision making, and evaluation and regulation needs. We intend to study the state of the art, identify future research challenges, and reach a consensus on practical way to address these challenges.
2. Build a community of researchers for the HUMAINT project to collaborate with.

## 1.1  Workshop details

The workshop was structured in a set of short position presentations followed by panel discussions. Presentation slides can be found at the workshop web page https://ec.europa.eu/jrc/communities/community/event/humaint-kick-workshop.

**DAY 1**

9.00-9.30: Welcome (Jutta Thielen-del-Pozo, Vanesa Daza, Emilia Gómez)

**(1) Human vs machine intelligence**

9:30-12:00: Presentations

- Joan Serrà. Unintuitive properties of deep neural networks.
- Gustavo Deco. Whole brain modelling and applications.
- Karina Vold. Extended minds and machines.
- Rubén Moreno-Bote. Slow and fast biases in decision making.

12:00-13:00: Panel discussion. Presentation and moderator: Ramón López de Mántaras

13:00-14:00: Lunch

**(2) Algorithms' impact on human behaviour**

14:00-16:30: Presentations
- Henk Scholten. Digital transformation and governance of societies.
- Nicole Dewandre. Artificial intelligence: an interesting leverage point to rethink humans' relations to machines…and to themselves.
- Carlos Castillo. Algorithmic bias.
- Fabien Giraldin. Experience Design in the Machine Learning Era.

16:30-17:30: Panel discussion. Presentation and moderator: Verónica Dahl

### DAY 2

**(3) Evaluation and regulation of algorithms**

9:00 - 11:30: Presentations
- Alessandro Annoni. Digital transformation and artificial intelligence: the policy-oriented perspective.
- Martha Larson. Reality, requirements, regulation: points of intersection with the machine learning pipeline.
- Anders Jonsson. Benchmarks and performance measures in artificial intelligence.
- Ansgar Koene. The IEEE P7003 Standard for Algorithmic Bias Considerations.
- Heike Schweitzer. Algorithmic decision-making - in need of (which) regulation?

11:30 - 12:30: Panel discussion. Presentation and moderator: Xavier Serra.

12:30-14:00: Lunch

**(4) Application domains and new paradigms**

14:00 - 15:00: Presentations
- Sergi Jordà. Enhancing or Mimicking Human [Musical] Creativity? The Bright and the Dark Sides of the Moon.
- Miguel Ángel González-Ballester. Machine learning in healthcare and computer-assisted treatment.

- Fabien Gouyon. The influence of machine intelligence on the music industry.
- Blagoj Delipetrev. HumanAI.
- Luc Steels. Will AI lead to digital immortality?

15:00 - 16:00: Panel discussion. Presentation and moderator: Perfecto Herrera.

16:00 - 17:00: Wrap-up session. Moderator: Emilia Gómez.

## 1.2 Goals and structure of the report

The goal of this report is to provide an interdisciplinary state of the art overview on the interaction between human and machine intelligence and identify which are the research challenges related to this interaction and the practical ways to address them.

In order to do so, the document is structured in five parts. The first four parts are related to the four sessions of our workshop. Each part contains a series of original contributions and position statements provided by workshop presenters. They are complementary to their presentation slides and contain their statements with respect to a set of questions proposed beforehand. In addition, each part includes a summary of the workshop discussions provided by panel moderators. In part V, the report incorporates input from other scholars involved in our discussions that were not able to present at the workshop.

The report finishes with some general conclusions of this interdisciplinary discussion and of the HUMAINT project and some directions for future research within the HUMAINT project and beyond.

## Part I: Human vs Machine Intelligence

In this part of the document, we address the following research questions:

— Which are the fundamental differences between human and machine intelligence?
— How do algorithms complement or replace human tasks now and how will they do this in the future?
— Will algorithms that take over some of our tasks affect the balance between human and machine intelligence?

We present a set of statements which address these questions from different disciplines and views, and provide a summary of discussions.

# 2 Unintuitive properties of deep neural networks

Joan Serrà

Telefónica Research

## 2.1 Statement

Deep neural networks are currently a hot topic, not only within both academia and industry, but also among society and the media. However, interestingly, the current success and practice of deep learning seems to be uncorrelated with its theoretical, more formal understanding. In particular, we find a number of unintuitive properties both in their design and operation that do not have yet an agreed explanation. Interestingly, some of these unintuitive properties are shared with humans, although current neural network approaches and the underlying mechanisms that lead to those properties do not resemble human mechanisms.

- **Neural networks can make dumb errors** — neural networks can produce totally unexpected outputs from inputs with perceptually-irrelevant changes, which are commonly called adversarial examples. Humans can be also confused by 'adversarial examples': we all have seen images that we guessed were something (or a part of something) and later we were told they were not. However, the point here is that human adversarial examples do not correspond to those of neural networks because, in the latter case, they can be perceptually the same (Szegedy et al., 2014).

- **The solution space is unknown** — as with many other machine learning algorithms, the training of neural networks proceeds by finding a combination of numbers, called network parameters or weights that yield the highest performance or, more properly, the minimum loss on some data. There are well known methodologies to find such a minimum for a few parameters with theoretical guarantees. However, deep neural networks are typically in the range of millions of parameters, for which a suitable combination that minimizes a certain loss must be found. The losses of current deep networks are non-convex, with multiple local minima and potentially many obstacles (Li et al., 2017).

- **Neural networks can easily memorize** — recent work empirically shows that finite-sized networks can model any finite-sized data set, even if this is made of shuffled data, random data, or random labels (Zhang et al., 2017). This has the obvious implication that neural networks can remember any data seen during training, no matter the nature of that data. What is not so obvious is that, still, if the data is not totally random, neural networks are totally capable of extrapolating their memories to unseen cases and generalize. Doing so when the number of model parameters is several orders of magnitude larger than the number of training instances is what is intriguing and contradicts conventional machine learning wisdom.

- **Neural networks can be compressed** — one can drastically reduce the number of parameters of a trained neural network and still maintain its performance on both seen and unseen data (Han et al., 2016). In some cases, the amount of pruning or compression is surprising: up to 100 times depending on the data set and network architecture. Besides practical considerations, the compressibility of networks poses several questions: Do we need a large network in the first place? Is there some architecture twist that combined with current minimum-finding algorithms allows to discover good parameter combinations for those small networks? Or is it just a matter of discovering new minimum-finding algorithms?

- **Learning is influenced by initialization and example order** — As with human learning, current network learning depends on the order in which we present the examples. Practitioners know that different sample orderings yield different performances and, in particular, that early examples have more influence on the final accuracy (Erhan et al., 2010). Furthermore, it is now a classic trick to pre-train a neural network in an unsupervised way or to transfer knowledge from a related task to benefit from additional sources. In addition, it is easy to show that random initializations can affect the final accuracy or, in the worst case, just prevent the network to learn at all.

- **Neural networks forget what they learn** — This phenomenon is known as catastrophic forgetting or catastrophic interference (McCloskey & Cohen, 1989). Essentially, when a neural network that has been trained for a certain task is reused for learning a new task, it completely forgets how to perform the former. Beyond the philosophical objective of mimicking human learning and whereas machines should be able to do so or not, the problem of catastrophic forgetting has important consequences for the current development of systems that consider a large number of (potentially multimodal) tasks, and for those which aim towards a more general concept of intelligence. Some research is devoted to tackle catastrophic forgetting, but a general solution for a compact model is still not yet fully in place (Serrà et al., 2018).

## 2.2 Challenges

- Adversarial examples are a big challenge right now. Perhaps we are not going to be able to solve the situation until some of the other the inner workings of neural networks are properly understood.
- Generalization is another principal hurdle. Current statistical theories for generalization are based on quite old models (like support vector machines and the like) that do not represent the state-of-the-art in many machine learning tasks.
- A flexible and straightforward solution to the problem of catastrophic forgetting, especially considering limited resources.

# 3 Whole brain dynamics and model

Gustavo Deco

Universitat Pompeu Fabra

## 3.1 Statement

**Whole-brain computational models** aim to balance between complexity and realism in order to describe the most important features of the brain in vivo. This balance is extremely difficult to achieve because of the astronomical number of neurons and the underspecified connectivity at the neural level. Thus, the most successful whole-brain computational models have taken their lead from statistical physics where it has been shown that macroscopic physical systems obey laws that are independent of their mesoscopic constituents. The emerging collective macroscopic behaviour of brain models has been shown to depend only weakly on individual neuron behaviour (Breakspear and Jirsa, 2007). Thus, these models typically use mesoscopic top-down approximations of brain complexity with dynamical networks of local brain area attractor networks. The simplest models use basic neural mass or mean-field models to capture changes in mean firing rate, while the most advanced models use a dynamic mean field model derived from a proper reduction of a detailed spiking neuron model [see (Cabral et al., 2017; Deco and Kringelbach, 2014) and references therein for a review].

The **link between anatomical structure and functional dynamics**, introduced more than a decade ago (Jirsa et al., 2002), is at the heart of whole-brain network models. Structural connectivity data on the millimetre scale can be obtained in vivo by diffusion weighted/tensor imaging (DWI/DTI) combined with probabilistic tractography. The global dynamics of the whole-brain model results from the mutual interactions of local node dynamics coupled through the underlying empirical anatomical structural connectivity matrix. The structural matrix denotes the density of fibres between a pair of cortical areas ascertained from DTI-based tractography. Typically, the temporal dynamics of local brain areas in these models is taken to be either asynchronous (spiking models or their respective mean-field reduction) or oscillatory (Deco and Kringelbach, 2014).

Adding the **temporal dimension** to standard FC analysis paves new ways to characterize the switching behavior of resting-state activity. However, the best methodology to assess it is still under debate. The most commonly used strategy has been to calculate successive FC(t) matrices using a sliding-window. Recurrent FC configurations are then captured by applying unsupervised clustering to all the FC(t)s obtained over time. However, the sliding-window approach has limitations associated to the window size, which affects the temporal resolution and statistical validation. Recently, new methods have been proposed to calculate the FC(t) at a quasi-instantaneous level, namely Phase Coherence Connectivity or Multiplication of Temporal Derivatives, which allow for a higher temporal resolution with the caveat of being more susceptible to high-frequency noise fluctuations. To overcome this issue, we hereby propose to focus on the dominant FC pattern captured by the leading eigenvector of BOLD phase coherence matrices. The key idea of this task is to focus on spatiotemporal dynamical biomarkers instead of the classical static grand averaged biomarkers (e.g. FC): In concrete, both during task and at-rest, we will identify whole brain dynamical micro brain states, by clustering the dominant dynamic functional connectivity (FC) patterns captured by the leading eigenvector of those matrices. Recurrent FC patterns – or micro states – will be detected and characterized in terms of lifetime, probability of occurrence and switching profiles in link with the subjects' performance on the Battery of behavioral tests, evolution of disease, and recovery.

**This type of whole-brain modeling could be used for crucial translational applications**. The basic idea here is to exhaustively stimulate off-line a realistic subject specific fitted whole-brain model in order to detect which type and locus of stimulation is more effective to reestablish a healthy dynamic of the whole brain (both under resting and task conditions) in order to expect that under that condition Hebbian learning will cause a meaningful recovery.

Thus, multimodal neuroimaging (DTI, fMRI) is essential for having patient specific tailored whole-brain models which can be studied exhaustively. Whole-brain models could be fitted in particular by the novel spatio-temporal dynamical features mentioned above which characterize the network dynamics in probability microstates space. In parallel, based on healthy control groups, we can characterize also those same features.

The idea is to discover, which kind of external stimulation (type and locus) would promote a transition from the patient specific affected probability microstate space to a healthy one. This study can be done exhaustively in off-line simulations. After that, one can try in vivo, with TMS, if those reestablishment of healthy spatio-temporal dynamics causes recovery.

# 4 Extended minds and machines

Karina Vold

Leverhulme Centre for the Future of Intelligence and Faculty of Philosophy, University of Cambridge

## 4.1 Statement

While there are have been many astonishing feats by AI in the last decade—even in just the last year—there continue to be many fundamental differences between human and machine intelligence. For one, many of the headline-making accomplishments by AI have been in highly specialized domains, or in what experts call Artificial Narrow Intelligence (ANI). Humans possess a more general kind of intelligence. We must, after all, perform a wide-range of tasks in order to successfully navigate our complex environments. Specialists are working towards building Artificial General Intelligence (AGI)—machines capable of skilled performance in a wide range activities. But there exist fundamental differences between human and machine intelligence that may complicate this project.

One difference that is often pointed to between humans and machines, especially amongst philosophers, is consciousness. There is *nothing it is like* to be a machine. Machines can be damaged, but they do not feel pain (Dehaene, Lau, and Kouider 2017). There is no uncontroversial scientific or philosophical theory of consciousness but, given the central role of phenomenology in human life and the success of our species, it seems at least prima facie plausible that consciousness has a cognitive function (for dissenting opinions see Chalmers 1996; Jackson 1982). In this case, without consciousness computers may never have human-like intelligence.

Another fundamental difference is that human intelligence has a long evolutionary history. The physical world has put many constraints on human intelligence: our brains need to be small enough to fit through a human birth canal and light enough to be carried around on our necks. Furthermore, our biochemical processing speeds are slow, running on less power than a refrigerator lightbulb. This might have been useful when we had to conserve enough energy to scavenge for food, to build shelters, and to procreate, but computers do not have to do any of these things. Machines already process information more quickly and efficiently than the human brain (Reardon 2018) and all of their resources can be expended on one, very narrow task—hence their success in specialized domains.

These differences raise questions about whether we could ever build humanlike intelligence, as well as why we should want to build machines that mimic our own intellectual constraints, especially if it is possible to bypass human intelligence entirely and leapfrog into what experts call 'superintelligence', which would surpass humans in many or all cognitive tasks.

Humans have done remarkably well considering the constraints on our biological bodies. An increasingly popular set of views in philosophy of mind and cognition maintains that we have achieved cognitive success by finding ways of moving our thinking outside of our bodies. We created language, for example, a complex representational system that enables us to communicate ideas and build on them over time. We also created technologies, from pens and paper to smartphones, which allow us to augment our biological capacities and simplify the cognitive tasks our brains need to complete (Clark and Chalmers 1998). This points to yet another difference between human intelligence and machine intelligence: our cognitive functions crucially depend on our bodies, our environments, and our tools. Our intelligence, as it is sometimes put, is embodied, embedded, and extended. To attain human-like intelligence, machines too might need to be embodied (perhaps even in human-like forms), embedded in their environments, and able to extend their cognitive functions beyond their hardware through seamless integration with tools.

One concern that falls out of this idea that machines may move beyond their intended, or original, hardware base to make use of other tools to complete their tasks is whether us humans will become their tools of choice. We have already seen instances where the outcomes of algorithms have played a role in shaping human decision-making, e.g. in the application of risk-assessment algorithms in parole decision (Kehl, Guo, and Kessler 2017). Ng (2016) argues that any mental task that a typical person can do in less than one second of thought can be automated, which suggests that the more difficult tasks to automate will be precisely those that require us to reflect. Tasks that we can do in less than one second do not require consciousness (Kahneman 2011). Human consciousness may need to play a key reflective role in complementing algorithms, but will have to do so while being guarded from the biases that influence our 'fasting thinking' systems.

## 4.2 Challenges

- Does it make sense to draw comparisons between human intelligence and machine intelligence?
- Can these comparisons mislead us and even harm us? How can they be used to help humanity?
- How is interaction with machines affecting human intelligence and cognitive capacities? Is it augmenting and enhancing our capacities (Savulescu and Bostrom 2009), is it simply changing our capacities (Carr 2010), or is it perhaps diminishing them?
- How can humans rely on the suggested outcomes of algorithms without being unduly influenced by them?
- How can we prevent humans from being used, nudged, or manipulated by machines? And when, if ever, is relinquishing control to machines for the best?

# 5 Slow and fast biases in decision making

Ruben Moreno-Bote

Universitat Pompeu Fabra

## 5.1 State of the art

The brain is the only known intelligent system in the whole universe. It consists of 100 billion neurons working together in intricate circuits to generate complex behaviour that allows adaptation and survival of the species. The study of the brain is important for clinical aspects, but also because it can inform and inspire new waves of AI. Also, knowing how it works will permit smoother interactions between humans and future 'intelligent' technologies.

An example of the benefit of studying the brain in AI is Deep Learning, which originated from early inspirations of theoreticians of the brain: the inspiration was that artificial neuronal networks consisting of interconnected non-linear units could represent increasingly complex 'abstract' variables. Thus, it is expected that new inspirations for AI will come from how the brain works.

It is interesting to observe that algorithms for AI could be designed to work at high performance in specific problems, but do not necessarily generalize well to new domains or even new datasets. However, this is part the strength of algorithmic AI, as it is possible to design systems that work very well under well-defined and constrained conditions. This is beyond the scope of human brains, which are not designed to work in repetitive and highly constrained or restricted conditions. Also, search algorithms are more efficient in some respects than the human brain, especially when dealing with vast amounts of data in digital format, but not necessarily so in other formats.

We think that it is important to understand human and complex animals' behaviour in situations in which there is no a priori reason to expect that human behaviour can be worse than the algorithms designed by AI. For instance, in perceptual decision-making tasks, over-trained animals are expected to perform the task very efficiently, as it is typically observed. However, we observe biases from previous trials that affect performance. The presence of these biases is informative about how artificial the task is, and how much it deviates from its natural setting in which it was designed to operate. By knowing these biases, one can better compare human vs machine performance in situations in which experimental conditions can deviate from naturalistic environments.

## 5.2 Challenges

- Define tasks in which a priori there is no reason that subjects can do it wrong, and yet it is shown that performance is far from optimal, or biases are shown, or wrong tendencies are shown.
- Complementary, design tasks that are naturalistic and in which human behaviour is highly adapted and close to optimal. Then test performance of AI algorithms in these naturalistic settings.
- How biases can be eradicated from behaviour? Can machines/algorithms help to correct for this? Recent work shows that blocking of certain areas improves performance, as if the knob for biases could be removed.

# 6 Human vs machine intelligence: an interdisciplinary discussion

Ramón López de Mántaras

IIIA (Artificial Intelligence Research Institute) of the CSIC (Spanish National Research Council)

The Panel "Human versus Machine Intelligence" addressed several points based on the four presentations that we had right before the panel. Namely those of Joan Serrà on "Unintuitive properties of Deep Neural Networks (DNNs)" (see Section 2), Gustavo Deco on "Whole brain modelling and applications", Karina Vold on "Extended minds and machines" (see Section 4) and "Rubén Moreno-Bote on "Slow and fast biases in decision making" (see Section 5).

The panel started with some initial remarks by the moderator briefly relating the four presentations. The main remark was to point out the **very big difference between the computational approaches based on deep neural networks and the real brain**. Indeed, from the presentations of Gustavo Deco and Joan Serrà it was pretty clear how complex the structure and the functioning of the brain is and how poor and limited are the artificial neural networks (ANNs) are, including DNNs. The practice in DNNs relies too heavily on "trial and error" and this is why we do not really know why sometimes they work so well whereas sometimes they make so dumb errors. The gap between theory and practice is very large! DNNs lack explanatory capabilities and they suffer from what is known as "catastrophic forgetting" which means that they forget the task they have just learned as soon as they are trained to perform a new task. This last fact in itself is an indicator of the big difference that exists nowadays between human and machine intelligence.

The presentation of Gustavo addressed the issue that the brain at "rest" (without stimuli) gives an output. So, the "resting" brain is in fact not resting and displays spatial patterns of correlated activity between different areas of the brain. In summary, from his talk, and the subsequent debate in the panel, we could see that the computational modelling of the whole brain dynamics is extremely difficult and extremely far away from the current computational models used in AI.

From the talk of Karina Vold and the panel debate we could see other **clear differences between human and machine intelligence**, namely: consciousness, evolutionary history, embodiment, situated cognition (as part of cognitive extension), and very importantly social intelligence.

Indeed, we, humans, are social agents and this fact obviously "shapes" and extends our intelligence. **Such fundamental differences raise serious questions regarding whether the goal of building humanlike intelligence is possible**. Another interesting aspect that was raised and discussed is that the technology we use can become **functionally integrated into our biological cognitive** capacities such that the tools become part of our minds on a pair with our brains (in the sense of what Clark and Chalmers call "extended minds") and how this can affect human intelligence and our cognitive capacities. Another related question is how we can prevent humans from being manipulated and if we should relinquish control to machines.

Finally, the presentation of Ruben Moreno-Diaz was about decision making in animals and machines and particularly on the presence of slow and fast biases in decision making due to previously existing preferences and information (slow versus fast biases distinction is related to how long before the previous preference was chosen or how long before we had the piece of information that affected our choice). Given that we almost never start from an indifference state. One important question is whether we can avoid biases. The answer seems to be no in general. However, the biases can vary in strength and can be partially controlled. The strength of the biases seems to be related to how artificial the task -that is the object of decision- is, that is how much it deviates from the natural setting it was designed to operate. One claim of this work is that thanks to the presence of biases we can better compare human with machine performance in situations where the experimental conditions deviate from natural environments.

As a final wrap up summary we could say that **the state of the art of machine intelligence is still extremely far from human intelligence** and it is very controversial whether, in spite of the recent results based on deep learning, there has been real scientific progress towards the extremely ambitious goal of achieving human-like AI. Another relevant question is: Do we really need it?

## Part II: Algorithms' impact on human behaviour

In this part, we address the following research questions:

— How do algorithms, when exploited in different applications, affect human cognitive capabilities?
— How do algorithms have the potential to modify the way humans make decisions based on them (e.g. influence of recommendations, personalization)?
— Which are the suitable strategies for effective human-algorithm interaction?

# 7 Artificial Intelligence: an interesting leverage point to rethink humans' relations to machines…and to themselves.

Nicole Dewandre

Joint Research Centre, European Commission

## 7.1 Statement

The understanding and effect of the expression "Artificial Intelligence" are strongly conditioned by the implicit assumption that intelligence is –ideally- THE specific human (male) feature. "Artificial Intelligence" is spoken of, either from a creator's perspective, i.e. with pride or fascination, or from a slave's perspective, i.e. with fear and resentment.

In my contribution, I shall challenge both the creator's and the slave's perspectives and invite to a more agnostic and human-centric approach to AI.

1. "Intelligence" is much more easily granted to artefacts than to humans. Ex1: A fridge is deemed to be smart when it sends a signal to inform about a lack of milk. When a man or a woman scrutinizes the fridge to check if there is milk, this is not considered a smart task. Ex2: Public lighting is deemed to be smart if it adapts to the type of user (pedestrian, car, bicycle). If a man or women was posted on each street to turn the lights on according to the type of user, this would not be qualified as a high-skilled job! In fact, artificial intelligence is granted to artefacts *reacting to* their environment, when human intelligence is, instead, granted to behaviours *gaming* the environment, in order to reach an objective or materialise an intention. Mere reactivity, when it comes to humans, is not considered as intelligence, but rather as weakness. This leads to rethinking human intelligence, and the role of intelligence in characterising humanness.

2. In some places, we need to be reminded we interact with humans and not with machines. For example, a sticker encouraging saying "Hi" before asking for a ticket reveals that the default solution might have become to get a ticket from a machine instead of from the hands of another man or woman. In the same vein, interacting on websites, humans are asked to demonstrate to machines that they are humans, for example, through CAPTCHA, to be able to pursue the interaction. This reveals a generalisation of the fact that human-machine relationships are more and more positioning machine in the active role and humans in the passive or reactive mode. This way to see things occults the fact that machines and artefacts are owned and developed by agents, corporate or humans. So, instead of considering the human-machine interactions, the focus should be on human (owner/developer)-machine-human (user) interactions. This leads to rethinking the radical changes in the way (smart) artefacts mediate human relations.

## 7.2 Position regarding the research questions

**How do algorithms have the potential to modify the way humans make decisions based on them (e.g. influence of recommendations, personalization)?**

As the way humans make decisions is highly conditioned by their environment, and the environment being more and more pervaded by connected artefacts and algorithms, it is obvious that algorithms will impact the way humans make decisions. I would be cautious using the term "modify" as if there was a "before" and an "after" algorithms in the way humans make decisions. This question has to be addressed with the baseline of actual decision-making (partly contingent, depending from partial information, based on some random or serendipity) and not idealised decision-making (based on perfect, unbiased information).

**Which are the suitable strategies for effective human-algorithm interaction?**

As suitable strategies might depend from addressing the users' perspective or the owner's perspective, I shall deliberately answer this question from the user's perspective. From the users' perspective, it is essential to protect the attentional sphere of the users. Potentially, connected machines and AI could cannibalise human attention to a point that human attention is totally "sucked up" by machines, and humans are prevented from directing their attention according to their own desire or to each other. I recommend focussing on the situation of users faced with multiple systems instead of thinking of each application separately. Besides protecting the vulnerability of our attentional spheres, it is also essential to re-create the conditions enabling trust, and making sure that fooling each other is not a winning strategy. Instead of addressing these issues through control, I would recommend a minima- recreating the conditions allowing each of us to know if he or she is in an environment which "recognizes" him or her, and what is the impact of this recognition. For example, when I look for a good or service on the internet, is the price which is offered the same that anybody else would be offered? And if not, what is the impact of the environment adapting to me (in this example, with dynamic pricing)?

## 7.3 Challenges

- Critical approach of intelligence: what it is; its role in characterising humanness; the articulation between intelligence, knowledge and power.
- What is the effect of AI on human relations, and on the relation between humans and their environment (mix of artefacts and nature)?
- What does AI *not* change? In other words, what are the invariants with the past?

# 8 Algorithmic discrimination

Carlos Castillo[1]

Universitat Pompeu Fabra

## 8.1 Statement

Statistical group discrimination is disadvantageous differential treatment against socially salient groups based on statistically relevant facts (Lieppert-Rasmussen, 2013). In this definition, a group is socially salient if membership is important to the structure of social interactions across a wide range of social contexts; this includes in particular categories that are protected by law, such as individuals with disabilities, and groups defined by properties that are a matter of anti-discrimination law, such as gender, age, religion, national origin, etc. Algorithms based on statistical learning can engage in statistical group discrimination, if we understand statistically relevant facts as any information derived from training data — indeed, there are many examples of this (Hajian et al. 2016). Hence, <u>the interaction between humans and algorithms should be one in which the human is able to not only understand but also challenge algorithmic decisions</u>. The "FAT" framework (Fairness, Accountability, and Transparency) has been advanced in recent years as a set of characteristics to which algorithms should adhere.

### Recommendations

- There is a need for more awareness of the limitations of algorithms and the extent to which they can do harm.

- There is a need for robust evaluation frameworks that do not stop at abstract evaluation metrics (such as accuracy and precision) but that consider other elements of how an algorithm can affect the lives of people.

- There is a need for mechanisms of effective algorithmic transparency, which is not mere access to source code but a degree of algorithmic explainability that enables humans to understand and challenge algorithmic decisions.

## 8.2 Context

Despite lacking "subjective" elements in their decisions, algorithms, particularly predictive modelling algorithms that are used to support decision-making, can be discriminatory. A more precise formulation follows.

**Generic discrimination**

This and following definitions are adapted from Lippert-Rasmussen [2013].

X <u>discriminates</u> against someone Y in relation to Z if:

1. Y has property P and Z does not have property P (or X believes Y has property P and X believes Z does not have property P)

2. X treats Y worse than s/he treats or would treat Z

3. It is because Y has P (or because X believes Y has P) and Z does not have P (or because X believes Z does not have P) that X treats Y worse than Z

In other words, generic discrimination is <u>disadvantageous differential treatment</u>.

**Group discrimination**

X group-discriminates against Y in relation to Z if:

1. X generically discriminates against Y in relation to Z

1 C. Castillo is partially funded by La Caixa project LCF/PR/PR16/11110009.

2. P is the property of belonging to a socially salient group
3. This makes people with P worse off relative to others

   or X is motivated by animosity towards people with P,

   or by the belief that people with P are inferior

   or should not intermingle with others

"A group is socially salient if perceived membership of it is important to the structure of social interactions across a wide range of social contexts" [Lippert-Rasmussen, 2013]

A socially salient group could be, for instance, gay people. A non-socially salient group could be, for instance, people with brown eyes.

**Statistical discrimination**

X statistically discriminates against Y in relation to Z if:

1. X group-discriminates against Y in relation to Z
2. P is statistically relevant (or X believes P is statistically relevant)

For example:

- If an employer does not hire a highly-qualified woman because among his/her current employees, women have a higher probability of taking parental leave, then this employer is engaging in statistical discrimination.
- However, if an employer does not hire a highly-qualified woman because she has informed him/her that she intends to have a child and take parental leave, then this employer is engaging in non-statistical discrimination.

**In statistical machine learning**

An algorithm developed through statistical machine learning can statistically discriminate if we:

1. Disregard intentions/animosity from the definition of group discrimination
2. Understand the "statistically relevant" part of the definition as any information derived from training data.

Below, we provide five examples of discrimination in algorithms.

Disparate impact. The model gives people with P a bad outcome more often, or in other terms, people with P experience a higher <u>risk</u>.

Let's assume the following table:

|  | benefit | | |
| --- | --- | --- | --- |
| group | denied | granted | |
| protected | $a$ | $b$ | $n_1$ |
| unprotected | $c$ | $d$ | $n_2$ |
| | $m_1$ | $m_2$ | $n$ |

Figure 1. Benefit for protected and unprotected groups.

Suppose:

"Protected group" = "people with disabilities"

"Benefit granted" = "getting a scholarship"

Intuitively, if:

a/n1, the risk that people with disabilities face of not getting a scholarship is much larger than c/n2, the risk that people without disabilities face of not getting a scholarship, then people with disabilities could claim they are being discriminated (see, e.g., Pedreschi et al. (2012))

**Directly and indirectly discriminatory rules**. A model associates attribute P to a bad outcome, or attribute Q which depends on P, to a bad outcome.

Directly discriminatory rules

Suppose that from a database of decisions made in the past, after applying an associations rule mining algorithm, we learn that gender = female ⇒ credit = no

P(gender=female, credit=no) / P(gender=female) > θ

This means we have found evidence of direct discrimination (Hajian et al. 2013).

**Lack of calibration**. The same output translates to different bad outcome probs. for P and not P.

A model lacks calibration if the probability of an actual bad outcome depends on the class and not only on the output.

For instance, a well-calibrated model for risk of recidivism should have the property that, for every recidivism score generated by the algorithm, the probability of recidivism for different groups if the same.

**Disparate mistreatment / Lack of equal opportunity**. The false positive rate of the bad outcome is higher for P than not P.

Suppose we have the following distributions of scores (taken from ProPublica's research on COMPAS in 2016, https://github.com/propublica/compas-analysis).

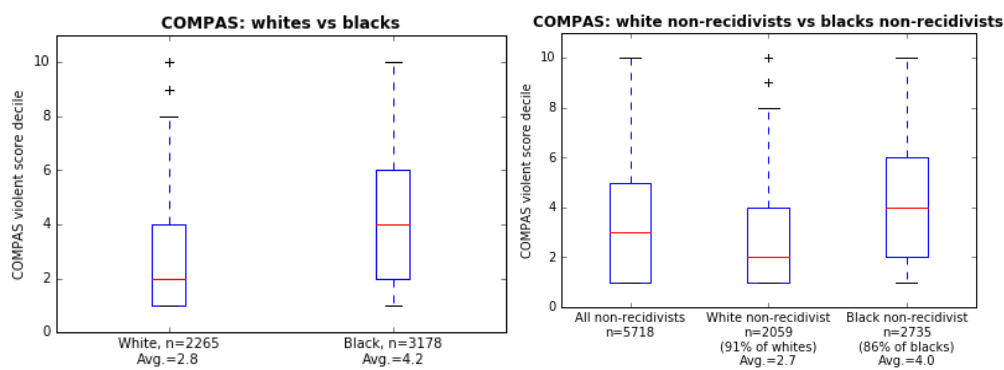

**Figure 2. COMPAS scores distribution.**

In the left plot, we see that the average score given to white and black defendants are different. This does not mean immediately that there is discrimination, but in the right plot, we observe that if we consider only those who did not commit a new crime (i.e., non-recidivists), the scores are still different. This gap means disparate mistreatment, or lack of equal opportunity.

**Unfair rankings**. The model gives people with P a lower ranking.

| Position<br>1 2 3 4 5 6 7 8 9 10 | top 10 male | top 10 female | top 40 male | top 40 female |
|---|---|---|---|---|
| f m m m m m m m m m | 90% | 10% | 73% | 27% |
| f m f f f f f m f f | 20% | 80% | 43% | 57% |
| m m m m m m f m m m | 90% | 10% | 73% | 27% |

Figure 3. Ranking comparison for different genres.

The results above are Top-10 results for job searches in XING (a recruitment site similar to LinkedIn), for selected professions: "Economist", "Market Analyst", and "Copywriter" (Zehlike et al., 2017). We observe that there is a difference in proportions between the top-10 and the top-40 in the three cases.

**Algorithm-human interaction**

Humans need to be able to receive explanations, and to correct outcomes.

Effective transparency does not mean source code, it means the human can understand and challenge the algorithmic decisions.

For instance, in RISCANVI (the method used to predict recidivism in Catalonia), experts can correct the assessment generated by the algorithm, indicating, for instance, that in their opinion the defendant has a higher/lower probability of recidivism that what the system generates.

## 8.3 Challenges

A personal opinion on transparency

- Many "customers" of algorithmic decision making systems do not value transparency. Transparency gives them insight on an algorithm and may generate doubts. Many want certainty, even if it is a false certainty.
- This is a perverse incentive for developers/providers, who may exaggerate their claims of accuracy.
- An additional problem is that numbers, plots, charts, suggest objectivity, so lack of mathematical literacy becomes problematic, as people tend to trust automated systems more than what they should.

Challenges

- **We need good evaluation frameworks.** How can this be fixed? Improving mathematic literacy and using evaluation frameworks that integrate multiple dimensions In addition to accuracy: "dollars saved, lives preserved, time conserved, effort reduced, quality of living increased" [Wagstaff 2012] and respect to privacy, fairness, accountability, transparency.
- To the extent that algorithms can engage in disadvantageous differential treatment that leaves people of a socially salient group worse-off, based on statistical information, **algorithms can discriminate**.
- Current research looks at trade-offs of utility and fairness and at mechanisms for mitigating unfairness.

## References

— Corbett-Davies, S., Pierson, E., Feller, A., Goel, S., & Huq, A. (2017). Algorithmic decision making and the cost of fairness. In Proc. of KDD.

# 9 Making quality decisions about the uses of algorithms and AI

Verónica Dahl

Simon Fraser University

## 9.1 Statement

**How do algorithms, when exploited in different applications, affect human cognitive capabilities?**

While algorithms can be very helpful as extensions of the human brain, unregulated algorithms can <u>tend to create cognitive dissonance</u> between our true power as humans and our perception of it, by giving us a disempowering sense of:

(a) <u>subordination to machines</u>: we must de facto submit to their modes of communicating, rather than vice-versa- e.g. call centers debug their defective speech recognition systems and decision trees for free at the public's temporal expense (an instance of the "time theft" crime), often with no helpful end results and no possibility to reach a human.

(b) <u>uncertainty</u>: decisions are often presented as unquestionable just because an impenetrable black box made them, leaving us unable to find out which criteria warranted a decision. This promotes either unwarranted blind trust or defeated resignation to uncertainty, not only among the general public but even in the very researchers developing or using the algorithms.

## How do algorithms have the potential to modify the way humans make decisions based on them (e.g. influence of recommendations, personalization)?

Algorithms being morally neutral, like tools are in general, they can either be used to create a happier state of the world, or to perpetuate and deepen inequalities and injustices, placing us at greater risk of global catastrophe. In Toby Walsh's words, the future is the product of the choices that we make today[2]. Given the speed at which AI advances unchecked and unregulated, we had better start foreseeing (and legally mandating) how algorithms and decisions based on, or made through them are bound to change the world, and how to best exploit them in order to gracefully adapt to the coming changes, that they may be for universal good.

Unfortunately, we are already cognitively prone to exhibit an unwarranted level of trust in algorithms when we make decisions based on them: we tend to treat information from an AI system as is it came from a trusted colleague, when in fact the system cannot even explain to us, like a colleague would, why it reached a given conclusion. A healthy skepticism is a must, as is the development of a proactive policy to address the potentially destructive consequences of algocracy (leaving decisions to machines), technological unemployment, and autonomous systems such as killer robots.

It is a fact of life that algorithms will be more and more capable of making decisions that were previously made by humans, and in general, of replacing humans, whose salary demands cannot compete with the under-regulated way in which algorithms have been allowed to replace them. But with enough foresight and legal provisions, we can revert this situation in ways that lead to humans universally getting to do work we value, and leaving the less desirable aspects of jobs to machines. As a scientific community we could gather around this and similar goals in order to prompt and help legislators in bringing the needed changes to fruition, and ensure that the results of research in AI can no longer be captured by a few powerful players to the detriment of the public that largely funded that research to begin with.

---

[2] Toby Walsh (2018) Machines that think: the future of Artificial Intelligence.

Regulation should also be put in place to prevent humans in power positions from making poor or abusive decisions just because these are now made possible by technology, e.g. automated tax reassessments can now be made in blanket fashion and be sent massively, creating a mountain of appeals that remain unsolved for years, since employees can't keep up with their processing. The consequences are dire for those who need tax clearance to proceed with their lives. Many end up simply paying, from either need of clearance or attrition, the undue amounts exacted.

## Which are the suitable strategies for effective human-algorithm interaction?

Effective beneficial human-algorithm interaction strategies require inclusive, democratic participation by humans in decision making. *Algorithms should be subordinated to humans*, serving them in such tasks as finding facts, proposing options and counting votes rather than making decisions that, because not collectively arrived at, cannot but be under-informed and biased.

Proportional gender representation is particularly key to intelligent decision making[3] and in general propitiates success for both sexes[4].

A good strategic bet for beneficial algorithmic interaction with humans might be to find and implement good automated methods for participatory and proportionally representative decision making, and adopt laws mandating governments to use these tools to obtain representative and specific mandates from the people as frequently as needed- at least for the most important questions that affect us all. The state-of-the-art in electronic voting allows us to ground democratic power in actual choice and consent. There is no excuse to continue the obsolete system of voting only once every X years for the blanket, static, and not even binding platform of some political party which often represents a minority when it wins.

This single strategy of algorithmically enabling and legally mandating proportionally representative decision making, if successful, could then serve to generate all other necessary strategies to ensure that algorithms serve all humanity, by democratic and representative, automated universal vote: Should flawed and/or unaccountable algorithms be allowed to replace humans? Should robo-signed mass actions against citizens be legal? Should algorithms pay taxes and benefits like the humans they purport to replace would? Should our laws allow for results of publicly-funded research to be used against the public, e.g. by creating unemployment? Should algorithms contribute to a fund for retraining humans into new jobs? Should our right to information be cancellable by the use of black box algorithms?

Even more importantly, this single strategy could also serve to generate the strategies that are urgently needed to push our Doomsday clock back from the two minutes to midnight it just hit: Should we all endorse the U.N.'s decision to ban nuclear weapons? What measures should be enacted towards ending violence, inequity, poverty, dominance, war, oppression, militarization, ecocide, climate change, etc?

We may not have the political will, as a society that has not quite reached true and representative democracy, to collectively and representatively induce the intelligent decision-making processes needed. But at least the necessary tools for reliably and efficiently mechanizing these processes are within our reach. As a community of scientists, we can, at this point of urgent need, decide to develop them and promote them into use in interaction with all other concerned groups besides ordinary citizens: scientists, educators, health care professionals, governmental agencies both national and international, legislators, judges, politicians, grass-root organizations, etc. Regardless of what means are chosen, our scientific community is already taking action, cf. their campaign urging a U.N. treaty against killer robots[5], or STEM professionals' initiative to more effectively manage technology and other resources crucial to human welfare[6].

---

3 https://futureoflife.org/2016/06/13/collective-intelligence-of-women-save-world/

4 https://news.ubc.ca/2014/09/30/gender-equality-olympics

---

5 https://www.theguardian.com/technology/2018/apr/09/killer-robots-pressure-builds-for-ban-as-governments-meet

6 http://demilitarize.org/milex-sign-new-statement-climate-change-military-spending/

# 10 Summary of the Panel Discussion "Algorithms' Impact on Human Behaviour"

| |
|---|
| Verónica Dahl |
| Simon Fraser University |

This discussion tied together Henk Scholten's talk on "Digital Transformation and Governance on Human Societies", Nicole Dewandre's on "AI as an interesting leverage point to rethink humans' relations to machines... and to themselves" (see Section 7); Carlos Castillo's on "Algorithmic Discrimination" (see Section 8); and Fabien Giraldin's on "Experience Design in the Machine Learning Era" (see Section 9). The moderator's brief position statement stressed as urgent the need to *regulate* algorithms, to ensure in particular that the wonderfully powerful tool that AI represents is used only for *beneficial* impact on human lives and behaviours.

The main questions discussed were:

- The need to debunk the view of rationality as the highest human capability, establishing in political terms the relational self as both free and social, so as to approach AI in a more human-centered (as opposed to control, malecentered) way.

- The need to systematically develop a proper vocabulary and mindset that will allow us to define what we need to do, for whom, and with appropriate measures of how, if adopted, it will lead to a better state of the world.

- The need to develop a vision of fairness in the digital world, and of what it means to evolve with AI in a socially-mindful, rather than interest-led, way.

- The need to correctly conceptualize notions of fairness and privacy, which are sometimes incorrectly invoked for the sake of the rational subject's interested wishes. While the separation line between using personal data for society's benefit and protecting it as private might be sometimes unclear, a good rule of thumb might be that someone's rights end where the rights of others begin, e.g. hiding behind encryption for criminal acts would warrant losing one's "right" to such privacy. Where the separation line is really blurry, laws designed for partial compliance (as exist already in Europe) are usually preferable to algorithmically enforcing total compliance.

- The need for transparency and accountability, defined as giving the public the ability to challenge an algorithm's decision (N.B. this is different from having all details about the algorithm, which may be useless in terms of challenging it). At present, Software Engineering cannot verify whether our complex, often statistically-based, unpredictably but speedily evolving AI systems will behave as planned, but perhaps a way will be found in the future. Meanwhile it may be prudent to legally disallow algorithms that cannot deliver transparency and accountability where it is due.

- The question of why are we being so suspicious or untrusting came up: it goes back to the degradation of human-to-human relationships under our globalized neoliberal economic politics, which normalizes human instrumentalization. For as long as "progress" means to use less people (improve labour productivity), and machines are being built to dispense with humans, it is "natural" for humans to end up wondering when machines will become "human". A way out might be to legislate that the uses of (publicly funded in particular) AI must benefit, not hinder, the public, e.g. by making algorithms (in fact, those operating them) pay taxes and benefits, and contribute also to a fund for retraining into new jobs the people they've disloged.

- The need to decide as a society, and legislate, who controls the results of AI, what is done in AI, and for whom. The absence of adequate laws for common good endangers modern political order, since a few very powerful global companies can profoundly influence many areas of daily life unchallenged, potentially generating informational dictatorships able to manipulate the behaviours of humans and organizations alike, and even to erode representative democracies and world peace.

Some possible strategies were also put forward:

- We should define machine intelligence not by what exceptional people can do, but through valuing what is widely shared.

- Protect the attentional sphere of the users from the demands of multiple systems.

- Educate governments and the public about AI's fallibility and limitations, e.g. neural net based systems cannot be totally autonomous given that they can have catastrophic failures embedded which cannot be anticipated until they occur.

- Conjure as much, and as representative, citizen participation as possible for decision-making on how to instrumentalize AI and technology in general for social benefit (e.g. [7]) . Many frameworks are possible, which must be balanced for efficiency and to not cause fatigue, e.g. as part of municipal bills of rights to be developed, or as machine assisted collective vote for the more critical issues (See more details in [8]). Whatever the framework, proportional representation stands out as the crucial ingredient for generating the most intelligent decisions[9].

- Work with legislators to bring about the needed laws that will protect us from the main dangers of unregulated algorithms (such as technological unemployment, algocracy, killer robots), and will ensure that they are used to help solve humanity's present problems for universal benefit rather than for that of a privileged few.

---

8 George Monbiot (2017) Out of the Wreckage. Verso.

9 https://futureoflife.org/2016/06/13/collective-intelligence-of-womensave-world

## Part III: Evaluation and regulation of algorithms

In this part, we address the following research questions:

— How should algorithms be evaluated in a research vs industrial context?
— Which are policy needs in terms of the usage of algorithms into real applications?
— Which is the research needed to support these policy needs?

# 11 Reality, requirements, regulation: Points of intersection with the machine-learning pipeline

Martha Larson

Radboud University

## 11.1 Statement

Intelligent systems such as search engines and recommender systems play an important role in mediating our consumption patterns and our decisions. We use such systems daily in order to find useful documents in the large amounts of content available online, and we perceive their ability to greatly exceed that of a single human searching by hand. Today's search engines and recommender systems go beyond documents such as webpages or books to also provide users with multimedia content (e.g., images, music, videos) and items, services, and opportunities in both the online and offline worlds (e.g., places to live, places to eat, jobs to apply for).

At the core of search engines and recommender systems lie machine-learning algorithms. These algorithms can be considered recipes that take a large amount of data as input and provide predictions (i.e., a list of results or recommendations) as output. In contrast to conventional computer programs, it is not possible to know exactly what the output will be in a given situation.

Search engines and recommender systems are considered intelligent since they produce in a blink of an eye a result that would have, in previous eras, taken a long consultation with a reference librarian, or a detailed discussion with the clerk of a video rental store. We experience these systems as doing something that humans are good at doing, doing it faster and doing it in an environment containing seemingly unlimited information. **However, how do we know that these systems are actually working the way we assume they are working?**

In order to answer this question, we must **evaluate the performance of AI systems**. Upon first consideration, it seems easy to argue that it is impossible to evaluate search engines and recommender systems. As stated above, the very reason why we build such systems is because we find human effort alone would fail to find the results that AI is capable of generating. If we as humans are not able to find the "right answer" to our information needs, given a large collection of documents, how can we possibly know that what an AI system finds for us is optimal, or even correct? The issue is compounded by the problem of not knowing in advance the output of an AI system in a particular situation, due to the nature of machine learning. However, the dangers are clear: we cannot base our decisions and behavior on systems we are not sure we can trust to be doing what they are supposed to do. Giving up on evaluating AI systems is not an option.

The way to proceed is to **return to engineering best practices**. In particular, we must specify a set of requirements for our systems, and verify that they meet those requirements. Ultimately, the position that it is impossible to evaluate search engines and recommender system is untenable because adequate effort has not yet been devoted to establishing requirements and to evaluating systems. The history of engineering is a history of finding solutions for problems that initially appear impossible.

In order to come near the amount of effort it will take to apply engineering best practices to the evaluation of search engines and recommender systems it is important to admit the possibility that AI evaluation is actually more difficult than AI itself: **If AI is smart, AI evaluation must be smarter.** Human engineers created AI in the first place. There is no a priori reason to assume that they are not also able to develop procedures and processes that are actually "smarter than AI". However, writing requirements and evaluating systems is a tedious and costly process, and must be recognized as such. If we are to insist on recommender systems and search engines that have been properly evaluated and demonstrated to be performing according to specifications, then we must necessarily give up our assumption that AI is some sort of a shortcut or is necessarily giving us something for free.

The return to engineering best practices will mean not just focusing on system output, but writing requirements for, and evaluating, every step along the machine learning pipeline (See Figure 4).

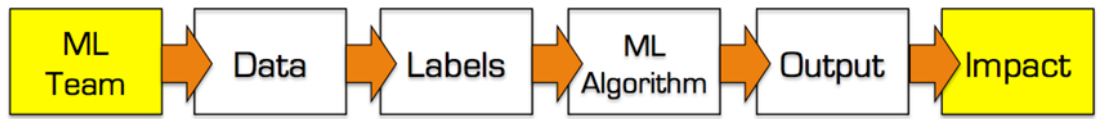

Figure 4. Machine Learning (ML) Pipeline. Evaluation of AI with respect to requirements and opportunities for regulation can be identified at every stage of the pipeline.

Formulating requirements for each stage of the pipeline requires considering the interplay of that stage with reality (i.e., with what the machine learning system attempts to capture). These stage-wise requirements provide detailed insight in what is happening inside an AI system. They can be seen as handles that allow us to get a grasp on the functioning of the system and to guide it toward desirable behavior and away from harmful behavior. The requirements represent an opportunity for regulation, which prevents machine learning from causing harm to individuals or society.

It is important to be aware that **requirement specifications are in the interest of industry**, and that **grounding regulations in requirements has potential to support compliance**. Companies are highly motivated to work according to engineering best practices. A well-specified set of requirements allows a company to focus on its core objectives, and be sure that the resources it is using are being directed to accomplish these objectives. The benefits of requirements to companies using recommender systems can be seen in concrete requirements documents, e.g., (CrowdRec project, 2014) and joint industry-academic visions on evaluation (Said et al., 2012). Requirements also allow companies to understand how they can adapt to regulations, including both achieving compliance and updating business models. For example, data regulations might make it difficult to compete with respect to collecting users' personal data ("Data" stage in **Error! Reference source not found.**), freeing the company from the need to devote esources to this area, and allowing it to shift its attention to competing with respect to algorithms ("ML Algorithm" stage in **Error! Reference source not found.**). The ontribution of data protection law to the quality of data science R&D has been pointed out by Mireille Hildebrandt (2017), who states, "By requiring the specification of one or more legitimate purposes by the controller, data protection law unwittingly contributes to sustainable research designs that have a bigger chance of making good sense of the data than sloppy exploratory research that covers its tracks in the name of experimentation and the freedom to tinker." Here, we point out that the positive impact of regulation promoting engineering best practices need not be unwitting.

Each stage of the machine-learning pipeline presents its own challenges, and for none of them is it easy to write requirements. For example, in the case of image search engines, training labels for image data ("Labels" stage) must incorporate the multiplicity of users' perspectives on images (Larson et al., 2014)(van Miltenburg et al., 2017). Further, requirements for the stages may interact. For example, minimizing training data ("Data" stage) might speed up algorithms ("ML algorithms" stage) and promote user privacy ("Impact" stage) (Larson et al., 2017).

Finally, we point out that "ML Team" and "Impact" are shown in **Error! Reference ource not found.** with a highlight intended to draw attention to the fact that machine learning **starts with people** (the people on the ML team that creates systems) and ultimately **impacts people**. Keeping people central in the ML pipeline requires investing effort into education. Education should include engineering best practices, but also encourage people to understand themselves and their personal worth in an era in which more and more of our lives are mediated by digital technology, cf. e.g., the thought of Jaron Lanier (2010). Engineering best practices have the potential to remind us that **AI may appear magic, but it is the magic of a magician: we enjoy the tricks, but realize that they do not change physical world as we know it.** With time and effort, we can also break AI down into its component stages, specify the requirements for each stage, and evaluate and regulate it using those requirements.

## 11.2 Challenges

- Putting as much effort into designing the requirements for AI, and evaluating AI as we put into creating AI in the first place: concretely, this effort includes building on existing benchmarking initiatives and starting new initiatives.

- Ensuring that the requirements for "Output" and "Impact" are consistent with fair treatment of user populations (bias elimination) and individuals (prohibiting harmful micro-targeting). This in turn means new requirements on "ML Algorithms", but also on "Data" and "Labels".

- Ensuring that the next generation of machine learning scientists is trained with the skills necessary to apply engineering best practices to AI evaluation, and are also motivated to do so.

# 12 Benchmarks and performance measures in artificial intelligence

Anders Jonsson

Universitat Pompeu Fabra

Artificial Intelligence (AI) is the field of computer science that studies the automatic generation of intelligent behaviour from a computational point of view. The term "intelligent behaviour" is usually defined in terms of how difficult it would be for a human to perform a given task (Russell, 2009). Computational problems that are historically considered part of AI include reasoning, knowledge discovery, planning, learning, natural language processing, perception and the ability to move and manipulate objects.

AI algorithms are mainly evaluated along two dimensions: 1) theoretical properties and performance guarantees; and 2) empirical performance.

Historically, the two dimensions carried similar weight, and most AI algorithms were published on the basis of little or no empirical support. This has changed dramatically in recent years, particularly with the advent of deep learning (Goodfellow et al., 2016). Today, empirical performance is a major factor in deciding whether a given AI algorithm is published, and theoretical analysis carries less weight. In fact, most deep learning algorithms come with no performance guarantees whatsoever.

Even though empirical performance is perhaps the most immediate measure of how well an AI algorithm works, theoretical analysis should not be overlooked as an evaluation metric. Performance guarantees come in many forms: an algorithm may eventually converge to the optimal performance level, or converge to a performance level that is within some bound of the optimal. Such an algorithm is guaranteed to always work well, no matter which task it is asked to solve. In contrast, an algorithm with no performance guarantees may perform very well on one task, but fail miserably on another.

An excessive focus on empirical performance provides researchers with strong incentives to boost the performance of their own algorithm relative to other algorithms, since this means their algorithm is more likely to be published. Researchers often make strong claims about their empirical results, such as having "solved" a particular domain or achieving human-level or superhuman-level performance. These claims should often be taken with a grain of salt, and independent verification and reproduction of empirical performance is becoming an increasingly vital task in order to establish the correctness of published results and determine whether they carry over to other domains.

Within the scope of empirical evaluation, there is also the question of precisely how the evaluation is carried out. In the case of stochastic algorithms that depend on random elements, a natural performance measure is the average performance across multiple trials, enhanced with a variance measure to test for robustness. However, researchers often publish the average of the K best trials, often without stating how many trials were carried out in total (Henderson et alt., 2017). The performance of deep learning algorithms is also highly dependent on other factors, such as the initial random seed, the values of hyperparameters, the network architecture, etc. Researchers often do not publish the details of how their algorithm was configured, making it harder to reproduce the work.

To apply AI algorithms in real-world domains it is necessary to establish much more stringent evaluation criteria. Ideally, these criteria should be adopted not only by researchers implementing AI algorithms in real-world domains, but by most or all researchers in AI. In addition, publishing source code and data would make independent verification and reproduction much easier. These measures would make published results much more trustworthy and make it easier to determine which algorithms hold the highest potential for real-world problems. Often, relatively basic statistics are sufficient to strengthen the aggregate results of multiple trials, compare different alternative algorithms, handle unbalanced datasets, etc.

Another measure that typically strengthens published results is to establish a set of benchmark problems for a given domain. Sometimes benchmarks are accompanied by competitions in which algorithms square off against each other on a subset of the benchmark problems. Benchmarks are normally available to the public, and their purpose is to provide a much more unbiased system for comparing AI algorithms. The more benchmark problems exist and the more diverse they are, the more difficult it becomes to artificially boost the performance of a given AI algorithm. If benchmarks reflect the difficulty present in real-world problems, the performance of an algorithm on the benchmarks should have a higher chance of carrying over to other, similar domains.

As an illustration of the importance of benchmark problems, consider the problem of classifying objects in images. State-of-the-art algorithms for image classification have been evaluated for many years as part of the ImageNet Large Scale Visual Recognition Challenge, or ILSVRC (Russakovsky et al., 2015). From 2010 to 2015, the error rate of the winning algorithm at ILSVRC decreased steadily from 30% to 5%, which is similar to the error rate observed in humans. Since the datasets used for the competition are large and contain a lot of labelled examples for training and evaluation, and since many experiments have been carried out with humans acting as the classifier, many researchers conclude that state-of-the-art algorithms have become competitive with humans.

Benchmarks are not without problems, however. Sometimes benchmarks do not accurately reflect possible real-world problems of a given domain, which may lead researchers in the wrong direction. Sometimes so much computational power is required that only a few select companies or organizations have the computational resources available to solve all benchmark problems. The prospect of winning a competition may also cause researchers to implement algorithms that do not really advance the state-of-the-art. A common example are portfolios that are optimized to select between the most successful existing algorithms by deciding for example how many seconds of computational time should be allocated to each algorithm.

Comparing the performance of AI algorithms with that of humans is not only of academic interest but effectively determines when it may be beneficial to replace human expertise with algorithms. An important aspect that affects the quality of such a comparison is how easy it is to measure success in a given problem. At least part of the reason for the popularity of AI in games is that success is very easy to measure. In real-world problems success may be much harder to measure, however. Eventually, as AI algorithms become better at high-level reasoning, new performance measures will likely be needed since success is not only measured by the performance on a single isolated task, but performance across tasks and how well the algorithm can integrate information from different tasks.

# 13 The IEEE P7003 Standard for Algorithmic Bias Considerations

| Ansgar Koene |
| University of Nottingham |

## 13.1 Statement

The rapid advance in the application of algorithmic decision making and machine learning methods to real-world applications, like screening of job applicate CVs, public-sector resource allocation (e.g. policing) and autonomous vehicles, with potentially significant impact on peoples' lives has generated an urgent need for practical guidelines and industry (self-)regulation in order to ensure that the highest standards of responsible conduct are applied as these powerful new algorithmic systems are developed and deployed.

A key challenge when it comes to the regulation of algorithmic decision making systems is that any evaluation of the bias/fairness of these system must take into account the inherently socio-technical context of how the system is (intended to be) used. When used for impactful decisions, the norms that an algorithmic system must obey are not just statistical, but also legal, moral and cultural [Dansk and London 2017].

In recognition of these challenges professional associations such as the ACM and the FAT/ML community have responded by publishing Principles for Algorithmic Accountability [ACM 2017, FATML] and a Social Impact assessment statement for Algorithms [FATML]. Around the same time the IEEE launched the IEEE Global Initiative on Ethics of Autonomous and Intelligence Systems which is developing a document [IEEE 2017] and a series of ethics based industry standards aimed at moving the discussion beyond statements of principles toward practical standards and policies.

## 13.2 Future challenges

- There is a need for more multidisciplinary coordinated thinking about the ways in which algorithms impact individuals and society.
- There is a need for clear assessment and certification regimes to communicate to users which algorithmic systems have implemented best practices for avoiding algorithmic bias.
- There is a need for research on effective benchmarking and impact assessment methods, especially regarding social impacts that go beyond statistical assessment of disparate outcomes.

# 14 Algorithms and markets: a need for regulation?

Heike Schweitzer

Freie Universität Berlin

## 14.1 Statement

Three main developments characterize the ongoing digitalization of the economy: (1) The increasing amount and importance of automatically generated data for the creation of new products and services and for the organization of all economic activities; (2) the development of ever more sophisticated algorithms to make economic use of that data; and (3) the rise of new business models based on data and its systematic analysis and use.

These three developments fundamentally change the way we communicate, interact socially and act (and are treated) on the market. As a consequence, the existing social and legal order has come under pressure: Its ability to ensure the fundamental values of our societies – among them private autonomy, privacy and a sophisticated control of private and public power – can no longer be taken for granted.

- As a reaction to the new explosion of data, data analysis capabilities – and possibly also "data concentration" – we have to rethink the informational order as it has existed in the past. This includes a rethinking of data protection rules as a new legal infrastructure for the functioning of markets, a rethinking of the role and responsibility of information intermediaries and the like.

- The ever more widespread use of "autonomous agents" challenges our concepts of agency, responsibility and liability. We have to rethink the rules on decision-making, discuss the legal demands we want to place on decision-making by autonomous agents and the preconditions for attributing such decisions to natural or legal persons.

- To what extent do we want to bind autonomous decision-making software to anti-discrimination rules and/or other ethical standards?

- If decision-making by self-learning algorithm resembles more some sort of "intuitive"/unconscious decision making than a conscious/rational decision-making (Mireille Hildebrandt): To what extent do we require a rational justification of decisions based on transparent criteria such that their fairness can be subjected to legal control? When / how do we want to impose such requirements on private actors, as opposed to state actors?

- We have to understand how the use of algorithms affects markets and analyze whether adjustments to the existing rules – in particular: data protection, fair trading and competition rules – are needed to safeguard their well-functioning and fairness.

To make things more complicated, these three challenges are not separate, but deeply interlinked. We have to understand and handle the interdependency of orders (information order, decision making systems, market order).

The goal must be a reconceptualization of these orders and their adaptation to the new challenges of a digital society and economy in a way that preserves the fundamental values that we continue to regard as the basis of our democratic societies and social market economies.

## 14.2 Research questions

The focus here is on how the use of algorithms affects markets and whether, due to such effects, new needs for regulations arise.

In the currently ongoing debates, the focus is on three potential risks associated with the exponential growth of the use of algorithms in the marketplace:

(a) Algorithms can increase market transparency – will they thereby facilitate coordination among suppliers as well as buyers? Under what conditions? Product homogeneity and a small number of actors in the market are familiar factors. But in an algorithm-driven market environment, do we have to expect collusion also where product heterogeneity, innovation and heterogeneity of preferences prevail, and where products and services become more and more individualised? Do we need to adjust existing competition rules to deal with the risk of "algorithmic collusion", namely abandon the distinction between "independent decision-making" and collusion, so as to cover parallel behavior also?

(b) Algorithms can facilitate price discrimination among different buyers, possibly even enabling some sellers to approximate perfect price discrimination. Is this a challenge to the well-functioning of markets that the law should address? Does it threaten to undermine the general trust in the fairness of market-functioning?

So far, the main task algorithms perform in the platform economy is to find the right match between heterogeneous products and heterogeneous preferences. In this regard, they "discriminate" according to preferences. There is a worry, though, that based on detailed personal profiles and an increased algorithmic understanding of "situational" power algorithms will offer the same products at different prices to different consumers. Will this do harm to the well-functioning of markets – despite the fact that, arguably, output will be maximized? What are the distributional consequences? What are the consequences for trust in markets? Can buyers self-defend against such uses of algorithms? Is there a risk of a costly "arms' race"? Is there a need for transparency (a duty to disclose arguably already follows from unfair trading law) or for more intrusive regulation?

Does competition law provide a layer of protection? In the "old world", active consumers were meant to provide protection also to the lazy consumers. In a world characterized by personalization, this may no longer hold. Should competition law react by narrowing market definitions and expanding its concept of market power to cases of situational power? Or should private law react by specifying its concept of "bonos mores"? Can data protection law contribute to the solution of the problem, and if so: how?

(c) Which principles apply to the use of algorithms by digital information intermediaries – and in particular: by digital information intermediaries with some degree of market power?

The explosion of information needs to a new importance of information intermediaries that ensure an efficient matching of parties. What is the effect of these intermediaries on market functioning? The intransparency of matching algorithms may significantly decrease the risk of collusion between sellers in markets – all the more, since information intermediaries should generally not be interested in such collusion. Also, the presence of information intermediaries should lower the risk of consumer exploitation – at least to the extent that they offer a meaningful product and price comparison. New risks can arise when intermediaries themselves possess some degree of market power – either on the business side or on the consumer side; and all the more, if the intermediaries are vertically integrated. Do we need a generalized principle of "digital intermediary neutrality" to be implemented into the relevant ranking algorithms? or a principle that outlaws the algorithmic priorization of vertically integrated offers? A fiduciary duty of personal butlers (as a specialized version of information intermediary / agent) vis-à-vis a consumer using it? Are digital information intermediaries under an obligation to explain the ranking of their offers – in order to effectively outlaw discrimination and/or self-priorization?

(d) Primarily, markets are meant to ensure an efficient allocation of resources. In order to perform this function, a degree of trust in the well-functioning and fairness of these markets must exist. In the presence of algorithm-driven markets: Do we need new rules for ensuring such trust? Do we need to expand the existing anti-discrimination rules as they apply also to algorithmic decision-making? Do we need new rules to avoid consumer exploitation in the light of potentially new degrees of information asymmetry?

# 15 Evaluation and regulation of algorithms: summary of discussions

Xavier Serra

Universitat Pompeu Fabra

The panel on Evaluation and regulation of algorithms included five presentations addressing different topics and addressing them from very different points of view. Alessandro Annoni, from the Joint Research Centre of the EU, talked about "Digital transformation and artificial intelligence: the policy-oriented perspective" where he explained the current and future policy EU initiatives related to AI from a regulation perspective. Martha Larson, from Radboud University and TU Delft, talked about "Reality, requirements, regulation: points of intersection with the machine learning pipeline" in which she emphasized the proper development of benchmarking for evaluating information retrieval systems (see Section 11). Anders Jonsson, from the UPF, talked about "Benchmarks and performance measures in artificial intelligence" in which he presented various approaches to benchmarking in AI from an academic perspective (see Section 12). Ansgar Koene, from the University of Nottingham, talked about "The IEEE P7003 Standard for Algorithmic Bias Considerations" in which he described the IEEE initiative on Ethics of Autonomous and Intelligent Systems and the different standards that are being developed under it (see Section 13). Finally, Heike Schweitzer, from Freie Universität Berlin, talked about "Algorithms and markets - a need for regulation??" in which she introduced a legal perspective by emphasizing the challenges that AI technologies bring to lawyers, presented in Section 15.

There was an initial list of questions that the organizers proposed for discussion: How should algorithms be evaluated in a research vs industrial context? Which are the policy needs in terms of the usage of algorithms into real applications? Which is the research needed to support these policy needs?. There were some relevant contributions to these questions during the talks, but there was no time for discussion to address them in depth. A much more extensive and focused discussion would be needed.

## Part IV: Application domains and new paradigms

In this part, we address the following research topics:

— Presentation of several application contexts where there is an interaction between human and machine intelligence and new future paradigms in computation.
— Presentation of research areas that will have a future impact on how we understand machine intelligence.

# 16 Machine learning in healthcare and computer-assisted treatment

Miguel-Ángel González-Ballester

ICREA and Universitat Pompeu Fabra

## 16.1 Statement

Machine learning is in a phase of renaissance that is transforming practices in multiple fields. Beyond the application of previously existing techniques, novel developments in big data and deep learning have transformed the landscape of available methodologies. Furthermore, many of these developments are spearheaded by industry, which is leading the application of machine learning to everyday products and services, not least in healthcare.

Image processing, in particular, has suffered a revolution in the last few years. Current methods based on deep learning clearly outperform the previous state of the art, and their extrapolation to medical image analysis has shown very promising results in i.e. lung cancer detection.

Computer aided diagnosis is also benefitting from recent developments in deep learning and machine intelligence, particularly in enabling the analysis of large, heterogeneous sources of patient data, such as genetic tests, blood and cell samples, imaging explorations and unstructured information from the clinical history of the patient. Furthermore, these tools are also being applied to study the aetiology of complex diseases, by finding patterns in large patient databases.

Despite these impressive initial success stories in medical image processing and computer-aided diagnosis, strong limitations have become apparent. Modern machine learning is predominantly based on "black box" approaches, failing to provide reasoned interpretations of the diagnoses they provide. Doctors (and we are far from replacing them) cannot afford to incorporate tools that provide a diagnosis with no explanation about the reasons behind this diagnosis. This poses a number of challenges for the widespread use of machine intelligence in healthcare.

## 16.2 Future challenges

- Interpretability is key to the future success of machine learning and artificial intelligence in healthcare. The combination of data-driven (empiricist) and model-based (Platonic) approaches might be key to this end.

- The availability of medical data is often limited by ethical and regulatory issues. Current trends in data augmentation and generative networks partly help in increasing the numbers of available data, but they risk biasing databases with unrealistic (non-disease related) information.

- Embodiment of artificial intelligence for patient care, e.g. through surgical robots, robot companions for the aged, or pervasive access and monitoring of health information, is an emerging discipline that may revolutionise healthcare.

# 17 The influence of "intelligent" technologies on the way we discover and experience music

Fabien Gouyon

Pandora

## 17.1 Statement

There is an influential loop between technological innovation, the development of business models governing the music industry, and the way we discover and experience music.

Internet and internet music streaming transformed the music industry. Streaming is now the primary way we listen to music, and this transformation is only at its beginning.

Until recently, the way we listened to recorded music was tightly linked to a relatively clear business model. Namely, music discovery would be driven by diverse media (e.g. terrestrial radio), targeting a subsequent purchase and ownership of a physical —or digital— artifact of what was discovered, consumption being done via another media (e.g. personal CD-player, iPod, etc.). A whole industry (a multi-billion dollar industry) was based on that model, which is now put under pressure.

Under this ownership model, once the discovery phase happens and an item is purchased, the job of content creators, producers and distributors is basically done. It is the listener who decides how and when to enjoy their music, with little influence from who produced or distributed it.

Now, with the advent of streaming, we are witnessing a shift from ownership to access. And with the access model, the line between discovery and consumption is now blurred, as the same media now serves both. There is an opportunity for content producers and distributors to guide listeners in their consumption. This opens the way to a much more holistic experience. And in return, the listener now requires to be assisted in all aspects from search, discovery, browsing, sorting through enormous collections of tracks, consumption, sharing, etc. That leaves room for many different novel models of listening experiences.

This is precisely where a crucial part of the music streaming industrial competition is currently happening. Diverse companies are developing at great speed new products, such as personalized playlists, radio-like lean-back propositions, etc., aiming at defining new formats of music listening. This calls for developing new technologies for recommending the most relevant content, as well as the most relevant vehicle for content discovery and consumption.

Such technologies must have the potential to be personalized for all listeners and reactive in real-time, and they must balance many factors such as e.g. content repetition, interactivity, or user intent.

In other words, the current developments in the music industry that are primarily driven by technological innovation are shaping the way hundredth of millions will access music, experience it and socialize around it. It is therefore fair to say that researchers and technological companies alike should acknowledge their strong cultural and societal responsibilities, and even further, embrace them.

Let's consider a few examples. For instance our responsibilities with the listener: The ubiquitous availability of (almost) any artefact of the world's music repertoire only a few clicks away imply overwhelming choices to music lovers. They need assistance, and we have a responsibility in helping them navigating through, and filtering this flood of content. Recommendation and personalization technologies can help in this endeavour. But they also are prone to potential algorithmic biases, and can result in a progressive isolation of users in their own musical bubbles, hence limiting and ultimately hurting their experience.

Let's now consider responsibilities with the music ecosystem, which we are part of: The fact is, music distribution is extremely unbalanced: a very small proportion of artists (the "head" of the distribution) account for most of what's listened to, while the large majority of artists (the "tail") remain listened by few. There are hits, and there are niches. We should carefully consider the effect technology can have on this distribution. It is (relatively) easy to develop technology that could —directly or indirectly— have a favourable distribution impact on either the "tail" or the "head". These would likely result in different economic returns for the industry and for the artists, on the short-term and on the long-term.

Finally, let's also consider that technological innovation is not only influencing the music listener experience as exemplified above, but it is in fact currently revolutionizing most aspects of the music industry, from creation, to rights monitoring, marketing, monetization, etc. This could spark similar reflexions on our cultural and societal responsibilities as developers of these technologies.

## 17.2 Challenges

- How, in the development of innovative technologies, can we exhaustively identify, understand and deal with algorithmic biases?

- How can we devise metrics that would approximate long-term user satisfaction?

- How can we make algorithms (e.g. recommendation algorithms) more transparent to their users? And what degree of transparency is actually desired/required?

- How can we balance ever more adaptive, contextual user experiences and respect for privacy? How can we provide users with more control on the data they provide us?

- What level of trust between a user and a technology/service is desirable to achieve? How far can the interaction go, and are there limits to be fixed?

- In the current context of extremely fast-paced scientific and technological developments, a very competitive and dynamic music streaming industry, and the "all-you-can-consume" nature of modern media, how can we help users cope with the overwhelming flood of content and products, and help them engage more deeply with content?

- What should be the basic methodological steps to follow so that the novel technology we develop not only follows to the latest technological trend, and responds to business metrics, but also empowers its users in its very evolution?

# 18 HumanAI

Blagoj Delipetrev

Digital Economy Unit, Joint Research Center, European Commission

## 18.1 Statement

**Human and Machine intelligence comparison**

There have been millions of year of biological evolution. Almost three billions years was needed for evolution to create the current human being. Evolution has increased human brain size exponentially over the last 8 million years from below 250 cc to 1500 cc. The brain is the source of our intelligence and separates us from all other species.

The industrial revolution, 2-3 centuries ago, produced machines that replaced physical human labour. The computational machines started 70 years ago with the semiconductors and their exponential grow which leaded to the creation of artificial intelligence (AI) which replaced humans in cognitive tasks.

The most important point is the **time scale**. Both human and machine evolutions are exponential and while human evolution is in millions of years, the machine evolution is in decades. There is a distinctive difference between human intelligence and the current AI, but this may not be the case in the future. Moore law is not dead, and the machine rise continues.

**Deep Learning**

Deep learning (DL) is the flagship of the AI research and achievements in the last decade. Increased computational power and vast amounts of digital data are foundation for DL, which is in essence a multiple layer neural network. DL achieved many breakthroughs, starting from vastly improving image recognition, NLP or translation. The DL combined with Reinforcement learning (RL) won the game of GO, Atari, Poker and lastly Dota. DL and RL are rapidly expanding.

**Algorithmic Impact Assessment**

The AI fast pace produced many systems that are used in everyday lives, government and public offices. There is a need of validation of these AI systems. Worldwide various initiatives for algorithm impact assessment will evaluate and analyse AI systems and their decisions and make them more transparent, understandable and explainable.

**Predictions**

My predictions are that "**everything electrified will be cognified**", "AI is the new electricity" making devices more intelligent and autonomous. Tasks described in productivity and efficiency will be performed by robots and bots. Now and in near future AI is going to complement us in all our daily tasks, as they do already, with our smartphone, computers, etc. In the middle term, the more advanced and intelligent machines will completely replace humans in tasks like transportation, medical image recognition, language translation, etc. In the long term is possible to have AGI or something close to it that will be capable of doing multiple cognitive tasks better than human does. This does not mean that humans will be obsolete.

In the meantime, we need to address many current problems and possible AI dangers. One of the most vivid dangers is autonomous weapons. Other highly important topics are the rising inequality, unemployment, fairness, inclusion, social justice, etc.

## 18.2 Challenges

The world is on the verge of one of its most valued discoveries AI. There are huge benefits in rising productivity, efficiency, improved standard, longer lifespan, better healthcare, etc. AI will automate most of the tasks, leaving more time for humans to enjoy lives and be creative. AI can bring more happiness and prosperity but also dead and destruction. Therefore, there is a need for AI regulation for the benefit of all humanity.

# 19 Do humans know which AI applications they do need?

Perfecto Herrera

Universitat Pompeu Fabra and Escola Superior de Música de Catalunya

In the last century, humans have developed powerful technologies that, for the first time in history have the potential to quickly and irreversibly change the world as it was previously known. Nuclear power and genetic engineering are application areas derived from useful essential knowledge (which cannot be questioned or censored) that had to be subject to ethical and legal regulation, even by means of international agreements. The current state of our knowledge on AI makes some of their applications to be about to cross (or maybe already crossing) red lines too and there have been attempts to reach a consensus, at least in the scientific community (see for example: https://futureoflife.org/ai-principles/ or http://www.iiia.csic.es/barcelonadeclaration/).

In the panel on the application domains of AI that closed our kickoff-meeting we witnessed some of such concerns, but also other ones that have to do with our concepts of humanity or creativity.

Sergi Jordà, in "Enhancing or mimicking human (musical) creativity? The bright and dark sides of the Moon" debated on attempts to develop "creative machines" and how many of them cannot shed enough light on human creative or other cognitive processes (one of the goals or justifications of some AI practitioners). In addition, creative systems are usually deprecated by their potential users (flesh-endowed music creators), not to mention the shallowness or uninterestingness of their outputs (though some outstanding exceptions could be considered). This rejection of creative systems could be due to a narrow developing perspective that does not consider the human to be inside a loop with the AI system. The idea of computers as assistants, becoming extensions of their users and doing the "dirty" or the short-time unfeasible work, should be promoted and researched (instead of leaving them the option to make the serious decisions). This is something that Luc Steels, another of the panel participants, commented during the panel dialogue ("Intelligence amplification" was the short-name given to that). A final issue with creative systems, but also with other AI devoted to "practical" problem-solving is that of understanding the outputs and the inner workings leading to them, from a human perspective, as some participants also remarked with the special case of game-playing AI systems.

The idea of assisting humans when dealing with creation is also challenging when they are not the creators but they enjoy an artistic creation, as Fabien Gouyon in "The influence of intelligent technologies on the way we discover and experience music" remarked (see Section 17). The almost permanent connection we listeners currently have with music, as a stream passing by or where you live immersed into, calls for ways to improve such listening experiences in "intelligent" ways (recommending truly relevant titles, helping to navigate through options, personalizing musical experiences, connecting with other people or groups, etc.).

Biases already noticed in other types of recommenders, and the potential construction of isolation bubbles on the side of users, should be carefully counter-acted in order to keep the collective and transformative power of music at its best.

A very different perspective and application field was discussed in "Machine learning in healthcare and computer-assisted treatment" by Miguel-Ángel González-Ballester (see Section 16), where advantages and shortcomings of current medical AI systems, especially those devoted to image-based diagnosis, were discussed. Here, again, the requirement that machine decisions can be interpretable under human (professional) criteria was raised. Additionally, the possibility that diagnosis might be done without humans in the loop (or that machines could bias or override the view and expertise brought by them) has also to be considered. An emergent topic, for which we are probably unprepared yet, was AI embodiment (i.e., what happens when humans incorporate, as body parts, AI systems?). Intersecting several of already mentioned hot issues, Blagoj Delipetrev, in "HumanAI How to assess algorithmic impact?" compared human and machine intelligence remarking what should not be considered as such (brute force approaches) and ways humans and machines could collaborate for a human-favourable scenario (see Section 18). He also discussed applications in satellite image recognition and on the assessment of the impact of AI algorithms. The necessity to deal with natural language and concept generalization before claiming "intelligence" for many AI systems was relevantly remarked there.

Our last talk, "Will AI lead to digital immortality?", by Luc Steels, speculated on future or "futuristic" applications and issues. Personal assistants are currently being developed under different appearances, and they will probably become autonomous artificial personae that might even impersonate different humans at the same time, not to mention that they might have multiple parallel lives, or that they will become immortal and then can continue with functions attributed to their formerly-assisted human beings. A list of ethical issues open by that perspective was barely touched, although, compared with those that the current intensively-used systems pose, they could be left for some future HUMAINT version 2.

During the open debate other important topics were lightly touched such as the apparent overabundance of AI-based start-up companies without clear business models (which can contribute to the "hype" of the topic), the risk of being monitored in concealed ways (by sound recording devices intended to play with toys or to just receive commands), the direct manipulation of behaviour that recommenders or notification services induce in their users, the blurring between what we thought reality is (or was) and what our senses are processing, the losing of some skills (some of them considered to be inherently human, such as caring for other beings, for example), or the consented (or worryingly ignored) externalization of some of our decision-making processes.

Applications of AI could sometimes be perceived as harmful because of their apparent extraordinary or superhuman "intelligence" but their most worrying aspects should be watched elsewhere: the difficulties to track or explain their decisions, the difficulties to embed AI systems with some moral sense or the subtle or blatant invasion of privacy they can facilitate. Even though there are enough examples of AI systems contributing to a healthier and more pleasurable existence (i.e., diagnosis systems, helpers for autistic or elderly people) and that such systems can help us to cope with data overload and strive in an increasingly complex reality, we should be watchful for some forthcoming large-scale disruption on the way we see ourselves and the surrounding world.

**It's time to decide how our future should look like, instead of leaving it to be devised just by what technology makes possible.**

## Part V: Considerations and conclusions

In this part, we first include a set of written contributions by other scholars that were not presented at the workshop but provided relevant input for our final considerations.

Then, we provide a set of conclusions to the workshop and directions for future work in the HUMAINT project.

# 20 Characterising the trajectories of artificial and natural intelligence

José Hernández-Orallo

Universitat Politècnica de València

Leverhulme Centre for the Future of Intelligence.

## 20.1 State of the art: An Atlas of Intelligence

The comparison between artificial and human intelligence is usually done in an informal and subjective way, often leading to contradicting assessments (Kirsh 1991, Hayles 1996, Brooks 1997, Pfeifer 2001, Shah et al 2016, Lake et al. 2017, Tegmark 2017, Marcus 2018). This is especially problematic because of the pace of the *epistemological* change. Our understanding of intelligence is rapidly progressing from new discoveries in comparative cognition, neuroscience and artificial intelligence. However, there is a possibly more relevant *ontological* change: artificial intelligence is creating new kinds of systems, and it is hence extending the landscape of intelligence (Sloman, 1984). Moreover, it is still not fully recognised ─and certainly not well understood─ that these technological changes are also affecting human cognition. Put it simply: because of AI, humans now think differently. Overall, and by all means, we have a moving target problem. Can we anticipate these trajectories?

In the first place, we need ways of assessing what AI systems can do, what they will be able to do in the near future, and the pathways and resources that will be needed to get there. Indeed, we need a common framework to determine which kinds of AI or hybrid systems in this landscape of intelligence are even desirable (needed for society) or possibly undesirable (dangerous, too similar to some profession profiles, animals or humans, etc.). For a recent symposium about this see: http://kindsofintelligence.org/.

The discussion must not be limited to the way society is affected: the irruption of AI systems with new capabilities may trigger a range of alterations in the very way human cognition works. The changes in memory capabilities, development trajectories and learning patterns that we are already observing because of the use of technology (negative Flynn effect, Google effect, etc.) can distinctly be regarded as cognitive atrophy or enhancement, but are about to change our psychometric profiles in possibly radical ways.

The Leverhulme Centre for the Future of Intelligence (http://lcfi.ac.uk) is working on a new initiative, an *atlas of intelligence*, to cover a relevant portion of the past, present and future landscape of intelligence: including humans, non-human animals, AI systems, hybrids and collectives thereof.

The atlas will be based on a set of dimensions, either behavioural features (i.e., the functionalities, cognitive abilities and personality traits) or physical features (i.e., the mechanisms, kinds of sensors and actuators, body morphology, computational or neurological resources). The atlas will allow users to make several projections and aggregations to a smaller number of dimensions. Also, despite the framework not being hierarchical, once populated, it could be converted into different kinds of taxonomies by using different distance/similarity metrics (as homology or analogy have been used for living systems) and also exploring a continuum from specialised (task-specific) systems to more general (task-independent) systems, including a developmental perspective.

The initiative is at an early stage and we welcome associates and contributors. More information about the specification and prospective maps to be considered for the atlas can be found in (Bhatnagar et al. 2017, 2018).

## 20.2 Challenges

- How can we characterise current and future AI systems in terms of cognitive abilities (Hernández-Orallo 2017a,b) and compare them to humans?
    - Are the new evaluation platforms (Castelvecchi 2016, Hernández-Orallo et al. 2017) going in the right direction? How do the AI milestones relate or compare to the milestones in child development or animal evolution?
    - Can we develop an ability-oriented analysis of job automation rather than task-oriented (Frey and Osborne 2017, Brynjolfsson and Mitchell 2017)?
- How can we characterise the changes in human cognition originating from technology and, most especially, from the interaction, replacement or enhancement with AI systems?
    - How can we analyse the locations and trajectories of human intelligence and AI progress?
    - How can the new "cognitive ecosystems" (Hutchins 2010), including humans and machines, be affected by these future changes of intelligence and their effect on dominance topologies (Cave 2017, de Weerd et al. 2017)?

# 21 Considerations related to cognitive development children

Nuria Sebastian

Universitat Pompeu Fabra

My comments will turn around two issues, both related to the concept of "development" that complements the current views.

## 21.1 Differences in machinery

There is a fundamental difference between "Artificial Intelligence" and "Human Intelligence" related to the enormous changes that the Human hardware (the brain) undergoes during life. Comparatively, artificial hardware undergoes relatively small changes (and the changes are not just in its size, but also in qualitative aspects).

The architecture of the human brain changes in fundamental dimensions during life. The most dramatic changes take place during the first years of life (I will not refer to prenatal changes / learning, because the point I want to make does not require to address this period, but there is learning during this time). The number of neurons increases exponentially mostly prenatally, but the number of synapses changes in a complex way after birth. It is worth noticing that changes do not take place in an uniform way across the brain. In general, sensory-related areas develop very quickly (reaching adult levels by the end of the first year of life), while "thinking-planning" areas (frontal areas) reach adult levels after puberty. On top of complex changes in the number of synapses, the amount of myelination (related to the effectiveness of neural transmission) diminishes with age (though it does not disappear).

These specific patterns impose important constraints the way the brain processes the inputs it receives. An important feature of brain development is the extraordinary synchronization between maturation of different brain areas: they become functional when they are needed. For instance: association areas become functional when "lower" areas are effective: there is no energy waste by having areas "waiting" for appropriate inputs. Functionality is the product of an exquisite interplay between internal development (gene-regulated, more prevalent early in life) and external input (more prevalent late in life).

On top of these substantial hardware changes, there are other important development-related changes in neurotransmitters and hormones that will have dramatic consequences on the way the brain functions across life. A clear case is the changes in sleep patterns taking place in life. Newborns spend most of their time sleeping, while elderly people tend to sleep very few hours. There are fundamental changes in the way the brain functions during sleep (not only quantitative, but also qualitatively) and it is well-known the critical importance of sleep in memory-consolidation.

Finally, newborns (from 4 hours to 4 days of age) and very young infants are able to perform complex computation over different types of signals (that seem to be relatively experience-independent). For instance, newborns can notice the difference between some (human) languages, such as Dutch and Japanese, if played forwards, but not backwards (as other species such as cotton-top tamarin monkeys and long evans rats). It is not until 5 months of age that human infants can distinguish English from Dutch, or Spanish from Catalan (it is worth noticing that by six months, infants already know several words). In a different domain, newborns prefer to orient to stimuli with a human face configuration than to a random one (Morton and Johnson, 1991).

In summary, there are essential differences between human and artificial "hardware" and they entail critical specificities regarding human learning.

## 21.2 Developmental changes in interaction with computers

The presentations have assumed that users interacting with AI systems are adults. However, there are important differences in the way children and adults deal with artificial systems, and this is an under-studied field.

One of the few existing studies has investigated the well-known Uncanny Valley effect. There is a vast literature investigating the fact that (human) adults feel uncomfortable when interacting with very human-like avatars/robots.

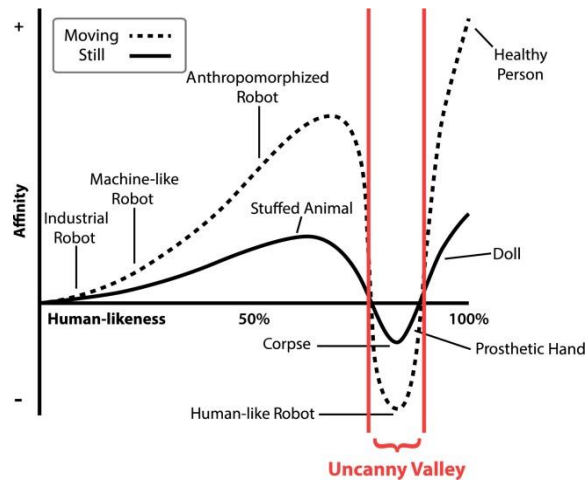

**Figure 5. Uncanny Valley effect.**

There is evidence indicating that such effect may be acquired. Young children (under 9 years) do not find "creepy" such very human-like avatars, importantly the feeling of "creepiness" is related to children's assumption that such avatars have human-like minds (Brink, Gray and Wellman, 2017).

The investigation of how children interact with machines is a virtually unexplored field. Such investigations are critical when considering the use of AI and robots in educational (and health) environments.

## References

— Kimberly A., Brink, K. G. and Wellman, H. M. (2017). Creepiness Creeps In: Uncanny Valley Feelings Are Acquired in Childhood. Child development, https://doi.org/10.1111/cdev.12999

— Morton, J., and Johnsson, M. H. (1991). CONSPEC and CONLERN: a two-process theory of infant face recognition. Psychol Rev. 1991 Apr;98(2):164-81.

# 22 The tyranny of data? The bright and dark sides of algorithmic decision making for public policy making

Nuria Oliver

Vodafone Research and Datapop alliance

## 22.1 Statement: Data-driven Algorithms for Public Policy Making

Today's vast and unprecedented availability of large-scale human behavioral data is profoundly changing the world we live in. Massive streams of data are available to train algorithms which, combined with increased analytical and technical capabilities, are enabling researchers, companies, governments and other public sector actors to resort to data-driven machine learning-based algorithms to tackle complex problems (Gillespie, 2014). Many decisions with significant individual and societal implications previously made by humans alone --often by experts-- are now made or assisted by algorithms, including hiring, lending (Khandani et al., 2010), policing (Wang et al., 2013), criminal sentencing (Barry-Jester et al., 2015), and stock trading. Data-driven algorithmic decision making may enhance overall government efficiency and public service delivery, by optimizing bureaucratic processes, providing real-time feedback and predicting outcomes (Sunstein, 2012). In a recent book with the evocative and provocative title ``Technocracy in America", international relations expert Parag Khanna argued that a data-driven direct technocracy is a superior alternative to today's (alleged) representative democracy, because it may dynamically capture the specific needs of the people while avoiding the distortions of elected representatives and corrupt middlemen (Khanna, 2017). Human decision making has often shown significant limitations and extreme bias in public policy, resulting in inefficient and/or unjust processes and outcomes (Fiske, 1998; Samuelson and Zeckhauser, 1998). The turn towards data-driven algorithms can be seen as a reflection of a demand for greater objectivity, evidence-based decision-making, and a better understanding of our individual and collective behaviors and needs.

At the same time, scholars and activists have pointed to a range of social, ethical and legal issues associated with algorithmic decision-making, including bias and discrimination (Barocas and Selbst , 2016; Ramirez et al., 2016) and lack of transparency and accountability (Citron and Pasquale, 2014; Pasquale, 2015; Zarsky, 2016). For example, Barocas and Selbst (2016) showed that the use of algorithmic decision making processes could result in disproportionate adverse outcomes for disadvantaged groups, in ways suggestive of *discrimination*. Algorithmic decisions can reproduce and magnify patterns of discrimination, due to decision makers' prejudices or reflect the biases present in the society. A recent study by ProPublica of the COMPAS Recidivism Algorithm (an algorithm used to inform criminal sentencing decisions by predicting recidivism) found that the algorithm was significantly more likely to label black defendants than white defendants, despite similar overall rates of prediction accuracy between the two groups (Angwin et al., 2016). Along this line, a nominee for the National Book Award, Cathy O'Neil's book, ``Weapons of Math Destruction", details several case studies on harms and risks to public accountability associated with big data-driven algorithmic decision-making, particularly in the areas of criminal justice and education.

In 2014, the White House released a report titled ``Big Data: Seizing opportunities, preserving values"[10] highlighting the discriminatory potential of Big Data, including how it could undermine longstanding civil rights protections governing the use of personal information for credit, education, health, safety, employment, etc.

For example, data-driven algorithmic decisions about applicants for jobs, schools or credit may be affected by hidden biases that tend to flag individuals from particular demographic groups as unfavorable for such opportunities. Such outcomes can be self-

---

[10]https://obamawhitehouse.archives.gov/sites/default/files/docs/20150204_Big_Data_Seizing_Opportunities_Preserving_Values_Memo.pdf

reinforcing, since systematically reducing individuals' access to credit, employment and education will worsen their situation, and play against them in future applications. For this reason, a subsequent White House report called for ``equal opportunity by design" as a guiding principle in those domains. Furthermore, the White House Office of Science and Technology Policy, in partnership with Microsoft Research and others, has co-hosted several public symposiums on the impacts and challenges of algorithms and Artificial Intelligence, specifically relating to social inequality, labor, healthcare and ethics[11]

At the heart of the matter is the fact that technology outpaces policy in most cases; here, governance mechanisms of algorithms have not kept pace with technological development. Several researchers have recently argued that current control frameworks are not adequate for situations in which a potentially unfair or incorrect decision is made by a computer (Barocas and Selbst, 2016).

Fortunately, there is increasing awareness of the detrimental effects of discriminatory biases and opacity of some data-driven algorithmic decision-making systems, and of the need to reduce or eliminate them. A number of research and advocacy initiatives are worth noting, including the Data Transparency Lab[12], a ``community of technologists, researchers, policymakers and industry representatives working to advance online personal data transparency through research and design", and the DARPA Explainable Artificial Intelligence (XAI) project[13]. A tutorial on the subject was held at the 2016 ACM Knowledge and Data Discovery conference (Hajian et al., 2016). Researchers from New York University's Information Law Institute --such as Helen Nissenbaum and Solon Barocas-- and Microsoft Research --such as Kate Crawford and Tarleton Gillespie-- have held several workshops and conferences these past few years on the ethical and legal challenges related to algorithmic governance and decision-making[14].

This chapter is a summary of the content discussed by Lepri *et al.* (2017a, 2017b), where the authors highlight the need for *social good decision-making algorithms* (*i.e.* algorithms strongly influencing decision-making and resource optimization of public goods, such as public health, safety, access to finance and fair employment) to provide transparency and accountability, to only use personal information --created, owned and controlled by individuals-- with explicit consent, to ensure that privacy is preserved when data is analyzed in aggregated and anonymized form, and to be tested and evaluated *in context* by means of living lab approaches involving citizens.

The opportunity to significantly improve the processes leading to decisions that affect millions of lives is huge. As researchers and citizens, I believe that we should not miss on this opportunity. Hence, I would like to encourage the larger community --researchers, practitioners, policy makers-- in a variety of fields --computer science, sociology, economics, ethics, law-- to join forces so we can address today's limitations in data-driven decision-making and contribute to fairer and more transparent decisions with clear accountability, within an ethical framework and developed by diverse teams so they can achieve significant positive impact.

## 22.2 Challenges

There are several limitations and risks in the use of data-driven predictive models informing decisions that might impact the daily lives of millions of people. Namely:

**21.2.1. Discrimination:** Algorithmic discrimination may arise from different sources. First, input data into algorithmic decisions may be poorly weighted, leading to disparate impact. For example, as a form of indirect discrimination, overemphasis of zip code within predictive policing algorithms can lead to the association of low-income African-American neighborhoods with areas of crime and as a result, the application of specific targeting based on group membership (Christin et al., 2015). Second, discrimination can occur from the decision to use an algorithm itself. Categorization can be considered as a form of direct discrimination, whereby algorithms are used for disparate treatment (Diakopoulos, 2015). Third, algorithms can lead to discrimination as a result of the misuse of certain models in different contexts (Calders and Zliobaite, 2013). Fourth, in a form of feedback loop, biased training data can be used both as evidence for the use of algorithms and as proof of their effectiveness (Calders and Zliobaite, 2013). The use of algorithmic data-driven decision processes may also result in individuals being denied opportunities based not on their own action but on the actions of others with whom they share some characteristics. For example, some credit card companies have lowered a customer's credit limit, not based on the customer's payment history, but rather based on analysis of other customers with a poor repayment history that had shopped at the same establishments where the customer had shopped (Ramirez et al., 2016). While several proposals have been made in the literature to tackle algorithmic discrimination and maximize fairness, we feel the urgency to establish a call for action bringing together researchers from different fields (including law, ethics, political philosophy and machine learning) to devise, evaluate and validate in the real-world alternative fairness metrics for different tasks. In addition to this empirical research, we believe it will be necessary to propose a modeling framework --supported by empirical evidence-- that would assist practitioners and policy makers in making decisions aided by algorithms that are maximally fair.

---

[11] https://www.whitehouse.gov/blog/2016/05/03/preparing-future-artificial-intelligence

[12] http://www.datatransparencylab.org/
[13] http://www.darpa.mil/program/explainable-artificial-intelligence
[14] http://www.law.nyu.edu/centers/ili/algorithmsconference

**21.2.2. Lack of Transparency/Opacity:** Transparency, which refers to the understandability of a specific model, can be a mechanism that facilitates accountability. More specifically, transparency can be considered at the level of the entire model, at the level of individual components (*e.g.* parameters), and at the level of a specific algorithm. In the strictest sense, a model is transparent if a person can contemplate the entire model at once. Thus, models should be characterized by low computational complexity. A second and less strict notion of transparency might be that each part of the model (*e.g.* each input, parameter, and computation) admits an intuitive explanation. A final notion of transparency might apply at the level of the algorithm, even without the ability to simulate an entire model or to intuit the meaning of its components. However, the ability to access and analyze behavioral data about customers and citizens on an unprecedented scale gives corporations and governments powerful means to reach and influence segments of the population through targeted marketing campaigns and social control strategies. In particular, we are witnessing an *information and knowledge asymmetry* situation where a powerful few have access and use resources and tools that the majority do not have access to, thus leading to an --or exacerbating the existing-- asymmetry of power between the state and big companies on one side and the people on the other side, conceptualized as a *"digital divide"* (Boyd and Crawford, 2012). In addition, the nature and use of various data-driven algorithms for social good, as well as the lack of computational or data literacy among citizens (Bhargava et al., 2015), makes algorithmic transparency difficult to generalize and accountability difficult to assess (Pasquale, 2015). Burrell (2016) has provided a useful framework to characterize three different types of opacity in algorithmic decision-making: (1) intentional opacity, whose objective is the protection of the intellectual property of the inventors of the algorithms. This type of opacity could be mitigated with legislation that would force decision-makers towards the use of open source systems. The new General Data Protection Regulations (GDPR) in the EU with a *"right to an explanation"* starting in May of 2018 is an example of such legislation. But powerful commercial and governmental interests will make it difficult to eliminate intentional opacity; (2) illiterate opacity, due to the fact that the vast majority of people lack the technical skills to understand the underpinnings of algorithms and machine learning models built from data. This kind of opacity might be attenuated with stronger education programs in computational thinking and *"algorithmic literacy"* and by enabling independent experts to advise those affected by algorithmic decision-making; and (3) intrinsic opacity, which arises by the nature of certain machine learning methods that are difficult to interpret (*e.g.* deep learning models). This opacity is well known in the machine learning community (usually referred to as the interpretability problem).

**21.2.3. Computational violations of privacy:** Reports and studies (Ramirez et al., 2016) have focused on the misuse of personal data disclosed by users and on the aggregation of data from different sources by entities playing as data brokers with direct implications in privacy. An often overlooked element is that the computational developments coupled with the availability of novel sources of behavioral data (*e.g.* social media data) now allow inferences about private information that may never have been disclosed. This element is essential to understand the issues raised by these algorithmic approaches, as has become apparent in the recent Facebook/Cambridge Analytica data scandal[15].

**21.2.4. Data Literacy:** It is of paramount importance that we devote resources to computational and data literacy programs aimed at all citizens, from children to the elderly. Otherwise, it will be very difficult, if not impossible, for us collectively as a society to make informed decisions about technologies that are not fully understood (Bhargava et al., 2015).

**21.2.5. Unclear accountability:** As more decisions that affect the lives of thousands of people are automatically made by algorithms, we need clarity on who is responsible for the decisions made by them or with algorithmic support. Transparency is generally thought as a key enabler of accountability. However, transparency and auditing do not necessarily suffice for accountability. In fact, in a recent paper Kroll *et al.* (2017) have introduced computational methods able to provide accountability even when some information is kept hidden.

**21.2.6. Lack of ethical frameworks:** Data-driven algorithmic decision-making poses important ethical dilemmas regarding what would be an appropriate course of action to take based on the inferences carried out by the algorithms or on the specific situation that the algorithm is acting upon. Hence, practitioners, developers, researchers and policy makers who would use data-driven algorithms to support or automatically make decisions would need to ensure that such decisions are made in accordance with a pre-defined and commonly accepted ethical framework. There are several examples of ethical principles proposed in the literature for this purpose[16][17] and institutes and research centers, such as the Digital Ethics Lab in Oxford or the AI Now Institute at NYU. However, it is an open question how to properly incorporate ethical principles in data-driven algorithmic decision making processes in addition to ensuring that all the developers and professionals involved comply with a clear Code of Conduct and Ethics.

**21.2.7. Lack of diversity:** Given the broad set of use cases that data-driven algorithms might be apply to**,** it is important to reflect on the diversity of the teams that generated such algorithms. To date, the development of the state-of-the-art data-driven, machine learning-based algorithms has been carried out by somewhat homogeneous groups of computer scientists. Moving forward, we need to ensure that the teams are diverse both in terms of areas of expertise and demographics –particularly gender.

---

[15] https://en.wikipedia.org/wiki/Facebook%E2%80%93Cambridge_Analytica_data_scandal
[16] https://www.wired.com/story/should-data-scientists-adhere-to-a-hippocratic-oath/
[17] https://futureoflife.org/ai-principles/

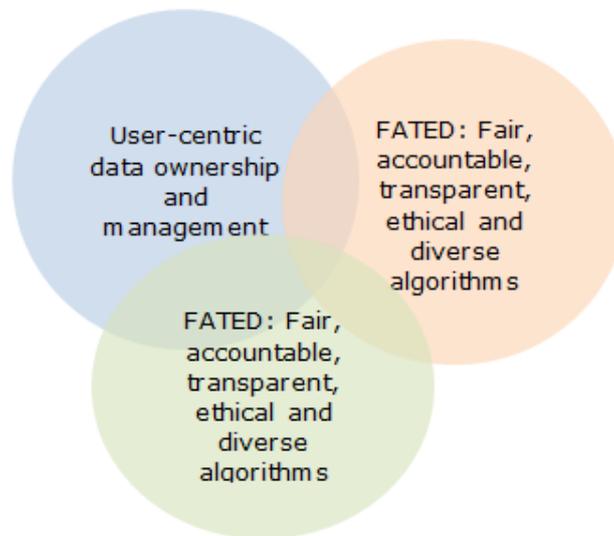

**Figure 6. Summary of requirements for positive data-driven disruption.**

While this is an exciting time for researchers and practitioners in this new field of computational social sciences, we need to be aware of the risks associated with these new approaches to decision making, including violation of privacy, lack of transparency and diversity, information and knowledge asymmetry, social exclusion and discrimination. I would like to highlight three human-centric requirements that we consider to be of paramount importance to enable positive disruption of data-driven policy-making: user-centric data ownership and management; algorithmic transparency and accountability; and living labs to experiment with data-driven policies in the wild. It will be only when we honor these requirements that we will be able to move from the feared tyranny of data and algorithms to a data-enabled model of democratic governance running against tyrants and autocrats, and for the people.

For the readers interested in the topic, they can find an extended version of this chapter in (Lepri et al., 2017a; Lepri et al., 2017b).

# 23 Quantum Computing and Machine Learning

Antonio Puertas Gallardo

Joint Research Centre, European Commission

Wide sectors of the industry and world economy are demanding more computing power and those needs are actually of a new king of computation. A growing request for High Performance Computing (or supercomputing) power exist amidst many areas (Finance, Chemistry, Pharma-industry, Nuclear fusion research), Big data and Artificial Intelligence, in general. The more digitalization of the economy increases, the higher is the request for a bigger and different type of supercomputing. Increasingly more systems and devices are clustering together, collecting data, in what is called the dawn of the Internet of things (IoT). Artificial intelligence processors are discovering mind-blowing levels of correlations or formulating inferences in huge amount of data, but still there are plenty of signals that a considerable numbers of companies are looking for new supercomputing paradigms. Classical computers (and supercomputer) are very big calculators that performed very well doing calculus and analytics using step-by-step operations, however quantum computing will be focused on the solution of problems from a more complex and higher point of view.

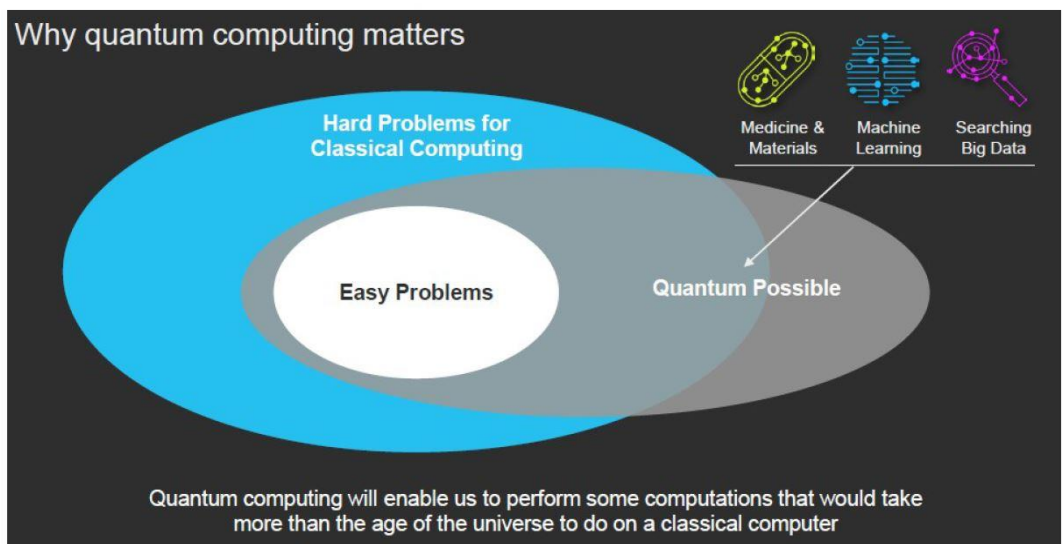

Figure 7. Areas of interest for Quantum Computing[18]

The capacity of data stored worldwide is increasing by 20 % on a yearly basis (nowadays is ranging in the order of hundredths of Exabytes) and is a compelling force to discover new approaches to Artificial Intelligence (Machine Learning). An encouraging new concept in computation is been now investigated by the most prominent IT companies research laboratories and Academic world, is the forthcoming and hypothetical utilization of quantum computing for the optimization of the algorithms of classic machine learning. Quantum computing will not render the classic computers inappropriate. Personal computers, notebooks and smartphones will still be running on silicon-processors for the likely future and the changeover may possibly take several years. Quantum computing might be the boosting element of the new "Fourth Industrial Revolution", likewise it might be, for example, a driver for the development of new molecules for drugs, the discovering of new materials and boost Machine learning algorithms that could not have been developed before with traditional computers.

---

[18] Source IBM

## 23.1 What is quantum computing?

A classical computer encodes information in the elementary unit of a logical bit, which can take values either "0" or "1", this information is stored and processed in the way of strings of bits (binary bits). Those individuals bit can have only one of two values: either 0 or 1. A quantum computer encodes information in the so called quantum bits or "qubits" each of them can simultaneously encode both logical bits "0" and "1" at once. This behaviour makes a quantum computer intrinsically parallel. The way to storage and process information in parallel, make some mathematical operations exponentially wide faster related to the computational speeds of classical computers for solving the same kind of problems.

Quantum computation exploits a quantum physics phenomenon called "superposition" which allows to a qubit (or quantum system in general) to be in a superposition of more than one state (not only "0" or "1" as conventional computers) at the same time. The differences between classical and quantum computers can be explained with the help of a coin. In classical computing, information is stored in bits with two states, either 0 or 1 – (or heads or tails). In quantum computing, information is stored in quantum bits ("qubits") that can be any state between 0 and 1 – similar to a spinning coin that can be both heads and tails at the same time. Among other advantages, a quantum computer makes computations by the manipulation of subatomic particles. These operations are faster and with lower energy consumption if compared with the classical computers.

Nowadays, the methods and instruments of quantum algorithms are very well founded and encompass a high amount of remarkable models and standards that overcome and beat the best established classical methods (See figure 4). The achievement of quantum computing is arising with IBM and Righetti succeeding in making their quantum computers available on the cloud. Many people are convinced that is only a matter of time until several theoretical designs can be tested on real-life machines. The innovative research discipline of Quantum machine Learning might offer the possibility to disrupt future approaches of intelligent data processing.

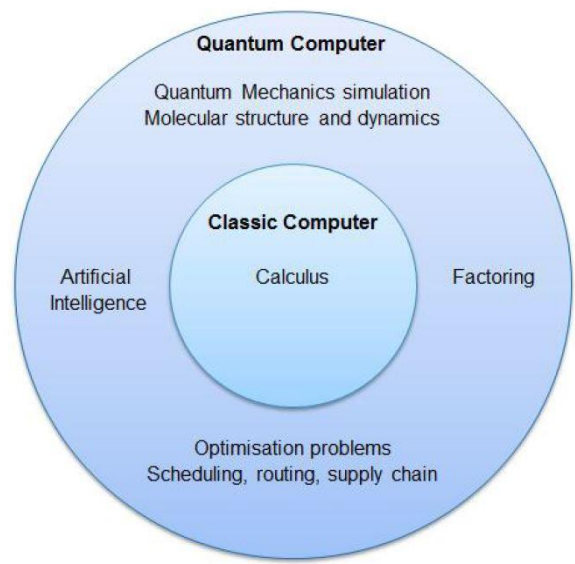

Figure 8. Computing science domains[19]

## 23.2 What is the holy grail of quantum computing?

**Exponential acceleration**

In other words, a quantum computer would be able to compute at a much faster speed (exponentially faster) than a classical computer. This implies that classical algorithms,

---
[19] Quantum computing – weird science or the next computing revolution Morgan Stanley Research Report, August 2017

which would take years to solve on a current supercomputer, could take just hours or minutes on a quantum computer.

## 23.3 Quantum Machine Learning

Quantum computation and quantum information have enabled us to think physically about computation, and this approach has yielded many new and exciting capabilities for information processing. Hence, it is possible to enable us to think physically (from a quantum physics point of view) about machine learning, especially about neural networks. The field of quantum machine learning explores how to devise and implement quantum software that could allow machine learning to perform faster than on classical computers. Quantum machine learning "QML" is the science and technology at the intersection of quantum information processing and machine learning.

To figure out the scientific research and work on quantum machine learning we need to consider it as a highway. On one side (one way), machine learning assists physicists to control and manipulate quantum effects and phenomena in labs. On the other side, quantum physics improves the implementation and performance of machine learning. In quantum machine learning, quantum algorithms are developed to solve typical problems of machine learning using the efficiency of quantum computing. This is usually done by adapting classical algorithms or their expensive subroutines to run on a potential quantum computer. The expectation is that in the near future, such machines will be commonly available for applications and can help to process the growing amounts of global information. The emerging field also includes approaches vice versa, namely well-established methods of machine learning that can help to extend and improve quantum information theory.

Quantum learning algorithms **have been realized in a host of experimental systems** and cover a range of applications as:

- Simultaneous spoken digit and speaker recognition and chaotic time-series prediction at data rates beyond a gigabyte per second (Brunner at al., 2013).
- Neural networks have been realized using liquid state nuclear magnetic resonance (Neigovzen et al., 2009).
- Defaulting on a chain of trapped ions, simulated a neural network with induced long range interactions (Pons et al., 2007).
- Solving a Higgs optimization problem with quantum annealing for machine learning (Mott et al., 2017).
- Quantum annealing versus classical machine learning applied to a simplified computational biology problem (Richard Y. Li, Rosa Di Felice et al, 2018)

## 23.4 Challenges

For the time being Quantum computing is still in transition between the Labs and the testing phase. This period is for the world of scientists and industry to focus on getting quantum-ready and to create a quantum-literate community who speaks quantum information language[20]

Artificial neural networks and machine learning have now reached a new era after several decades of improvement where applications are to explode in many fields of science, industry, and technology. The Emergent Quantum information technologies would eventually boost the impact on Artificial Intelligence.

1. - Machine learning algorithms training times could be accelerated exponentially[21].

2. - Parallelization of codes would be the new normal.

---

[20] https://www.symmetrymagazine.org/article/learning-to-speak-quantum

[21] https://www.youtube.com/watch?v=Q4xBlSi_fOs

3. - Software development would be revolutionized as programmers should need to learn to make codes which manage all solutions at the same time (instantaneously, when the algorithms are deployed into the Hardware layer).

# 24 Conclusions and future work

Emilia Gómez, Vicky Charisi, Bertin Martens, Marius Miron, Songül Tolan

Joint Research Centre, European Commission

This report has summarized the content of the 1st workshop on Human Behaviour and Machine Intelligence (HUMAINT), which provides an interdisciplinary view on the main challenges related to the study of the impact that machine intelligence will have on human behaviour and potential needs for policy intervention.

During the workshop, we have identified several research challenges and directions that can be summarized in the following ten points:

1. There are many **fundamental differences** between human and machine intelligence: consciousness, evolutionary history, embodiment, situated cognition and social intelligence. In fact, we often lack of a critical approach of intelligence: what it is, its role in characterising humanness, the articulation between intelligence, knowledge and power. For instance, intelligence is much more easily granted to machines than to humans in the current media landscape. Although there are major scientific advancements on the human brain and its computational modelling, this is an extremely complex endeavour and still far from current computational models used in AI application.

2. There is not yet a full **understanding** of the inner workings of **state-of-the-art deep neural networks**. As a consequence, estimation errors might be **unintuitive** for humans and **generalization** capabilities cannot be assessed. This limits the scientific understanding of algorithms, the capability to recover from adversarial examples, and complicates human supervision in practical applications. It also raises serious questions regarding whether the goal of building humanlike intelligence is possible and desirable. We should **monitor AI advancements** and new computing paradigms (e.g. quantum computing).

3. We need ways to **evaluate** what AI can do today and predict its potential future capabilities. We need to define evaluation frameworks that are meaningful and in naturalistic settings to match practical application contexts. In this respect, we should consider engineering best practices, impact assessment methods, user satisfaction and business metrics, in order to develop **smart and transparent benchmarking strategies**. We should train the next generation of machine learning developers to apply and communicate these strategies and follow **best practices** to AI evaluation.

4. We need to advance on the **explainability, accountability** and **transparency** of algorithms in general and deep learning architectures in particular, both from a machine learning research perspective (including theoretical understanding and empirical evaluation) and from a user perspective, when these methods are exploited in a particular application context. Humans should develop a critical thinking with respect to machine intelligence, and in order to do that people need to achieve **data and algorithm literacy**, so that everyone can understand and **challenge** it.

5. With respect to human vs machine intelligence, we should move from a competition to an **interaction** paradigm where we should research on best strategies for **collaboration and synergies exploitation between both intelligences**. For instance, we need to investigate on how biases can be identified in human behaviour and if algorithms could help to recover and correct this. Here we need to consider human(owner/developer)-machine-human(user) interactions, since **machine intelligence is in fact a product of human intelligence**.

6. It is important to understand the interaction in the context of **decision making**, e.g. considering **bias** present in algorithms and humans and how machines can be used to overcome human bias rather than incorporate it. This is particularly relevant in domains where decision making affects human welfare, e.g. in recruitment processes or the allocation of public funds. Also, we need to address how machines can affect **human attention** and strategies for humans to **trust** machines.

7. We need to research on how the interaction with machines **affects human intelligence and cognitive capacities**, if changing or diminishing them. In addition, we should research on how artificial intelligence changes relations between humans and between humans and the environment. While recent literature is focusing on the interaction between AI systems and adults, there are important differences in the way **children** deal with artificial systems that should be further researched, being the next generation to come.

8. Machine intelligence systems should be developed by humans in a responsible way. We should formalize and incorporate **ethical principles** in machine intelligence development and evaluation. We should also foster **diversity** (in terms of expertise and demographics, particularly gender), in teams that develop and are empowered with artificial intelligence to reflect varied perspectives into the developed systems.

9. Machine intelligence has a wide range of potential **economic implications**. There are already major concerns about the impact on human employment, wages and income distribution. The growing information asymmetry between humans and intelligent machines, and the potential for moral hazard and exploitation of human cognitive biases, will affect human behaviour and welfare. Competition between machines with scalable information processing capacities and humans with limited capacities will induce systemic shifts, including in the institutional structures of human societies.

10. There is a need to understand **who controls the results of AI**, what is done in AI, and for whom, and establish adequate **forecasting, control mechanisms and legal provisions** to anticipate and revert situations in which algorithms can be used against people's welfare, as well as establish adequate laws that ensure algorithms are used *for* people's welfare.

Finally, these mentioned points can be applied in different application domains. In this respect, we concluded that there are some aspects of algorithms that can be **analysed independently of the application context.** For instance, we agreed that the potential construction of isolation bubbles on the side of users in music recommender systems should be carefully counter-acted in order to keep the collective and transformative power of music at its best. This is also shared in other domains.

However, there are some other issues that should be considered **for particular use cases**. In terms of algorithmic transparency, for instance, the interpretability of algorithms is crucial in healthcare applications, while it can be less critical in music recommendation.

All of this requires **multidisciplinary thinking, diverse teams and future impact assessment**, as we should be watchful for forthcoming disruption on the way we see ourselves and the surrounding world.

# 25 List of workshop participants and report contributors

- Dr. Alessandro Annoni, Head of Digital Economy Unit, Leader of Artificial Intelligence Project.
- Dr. Benito Arruñada, Economy and Business Department, Universitat Pompeu Fabra, Barcelona
- Dr. Xerxes D. Arsiwalla, Institute for Bioengineering of Catalonia.
- Dr. Carlos Castillo, Social Computing and Web Mining.
- Dr. Vicky Charisi, Centre for Advanced Studies, Joint Research Centre, European Commission.
- Emanuele Cuccillato, Behavioural Insights and Design for Policy Unit, Joint Research Centre, European Commission.
- Prof. Veronica Dahl, Professor of Computing Science, Simon Fraser University, Canada.
- Prof. Dr. Gustavo Deco, Computational neuroscience, Director of the Centre for Brain and Cognition, Universitat Pompeu Fabra.
- Dr. Blagoj Delipetrev, Digital Economy Unit, Joint Research Centre, European Commission.
- Dr. Paul Desruelle, Digital Economy Unit, Joint Research Centre, European Commission.
- Nicole Dewandre, Joint Research Centre, European Commission.
- Dr. Fabien Giraldin (PhD), BBVA Data & Analytics.
- Dr. Emilia Gómez, Joint Research Centre, European Commission and Universitat Pompeu Fabra.
- Prof. Dr. Miguel Ángel González-Ballester, Simulation, Imaging and Modelling for Biomedical Systems, ICREA and Universitat Pompeu Fabra.
- Dr. Fabien Gouyon, Principal researcher, Pandora.
- Dr. José Hernández-Orallo, Universidad Politécnica de Valencia - Leverhulme Centre for the Future of Intelligence, University of Cambridge.
- Perfecto Herrera, Music Technology Group - Universitat Pompeu Fabra / Escola Superior de Música de Catalunya, Barcelona, Spain.
- Maria Iglesias, Legal Officer, Intellectual Property and Technology Transfer, Joint Research Centre, European Commission.
- Dr. Lorena Jaume-Palasi, co-founder and executive director of AlgorithmWatch (Germany).
- Dr. Anders Jonsson, Artificial Intelligence and Machine Learning group, Universitat Pompeu Fabra.
- Dr. Sergi Jordà, Music Technology Group, Universitat Pompeu Fabra.
- Dr. Ansgar Koene, University of Nottingham.
- Dr. Martha Larson, Radboud University and TU Delft.
- Prof. Ramón López de Mantaras, Director of the IIIA (Artificial Intelligence Research Institute) of the CSIC (Spanish National Research Council), Barcelona, Spain.
- Dr. Bertin Martens, Senior Scientist, Digital Economy Unit.
- Ever Meijer, Geodan.

- Dr. Marius Miron, Centre for Advanced Studies, Joint Research Centre, European Commission.
- Dr. Rubén Moreno-Bote, Centre for Brain and Cognition, DTIC, Universitat Pompeu Fabra.
- Dr. Pablo Noriega, Artificial Intelligence Institute, Spanish Council for Scientific Research (CSIC).
- Dr. Nuria Oliver, Vodafone Research and Data-Pop Alliance.
- Antonio Puertas Gallardo, Knowledge for Health and Consumer Safety Unit, Joint Research Centre, European Commission.
- Jordi Pons, Universitat Pompeu Fabra.
- Aurelio Ruiz, Universitat Pompeu Fabra.
- Prof. Dr. Heike Schweitzer, Free University of Berlin, Germany.
- Prof. Dr. Nuria Sebastian-Galles, Speech Acquisition and Perception Group, Centre for Brain and Cognition, Universitat Pompeu Fabra.
- Dr. Joan Serrà, Telefònica Research, Barcelona, Spain.
- Prof. Dr. Xavier Serra, Music Technology Group - Maria de Maeztu Strategic Program on Data-Driven Knowledge Extraction, Department of Information and Communication Technologies, Universitat Pompeu Fabra.
- Prof. Dr. Luc Steels, ICREA and Universitat Pompeu Fabra.
- Dr. Jutta Thielen del Pozo, Head of Scientific Development Unit and Director of the Centre for Advanced Studies, Joint Research Centre, European Commission.
- Dr. Songul Tolan, Centre for Advanced Studies, Joint Research Centre, European Commission.
- Dr. Karina Vold, Leverhulme Centre for the Future of Intelligence, University of Cambridge.
- Prof. Henk Scholten, University of Amsterdam, Co-Lead Scientist of Digital Transformation - Governance Project.

## List of figures